\documentclass[10pt,twocolumn,letterpaper]{article}

\usepackage{cvpr}              %

\usepackage{caption}
\usepackage{graphicx} 
\usepackage{colortbl}
\usepackage{graphicx}
\usepackage{multicol}
\usepackage{multirow}
\usepackage{makecell}
\usepackage{tablefootnote}
\usepackage{amsmath}
\usepackage{amssymb}
\usepackage{tabularx}
\usepackage[dvipsnames]{xcolor}
\usepackage{subcaption}

\usepackage{siunitx} %
\usepackage{colortbl}
\usepackage{graphicx}
\usepackage{lipsum}
\usepackage{nicefrac}
\usepackage{tabularray}
\usepackage{xspace}
\usepackage{pifont}

\usepackage{algorithm}%
\usepackage{algorithmicx}%
\usepackage{algpseudocode}
\usepackage[export]{adjustbox} %

\usepackage[accsupp]{axessibility}  %

\newcommand{\ours}{Chorus\xspace}

\def\eg{\emph{e.g.}\xspace} 
\def\ie{\emph{i.e.}\xspace} 

\def\wrt{\emph{w.r.t.}\xspace}

\def\scan{ScanNet\xspace}
\def\scannet{ScanNet\xspace}
\def\ppv2{ScanNet++\xspace}

\def\matt{Matterport3D\xspace}
\def\interior{InteriorGS\xspace}

\newcolumntype{R}{>{\raggedleft\arraybackslash}X} %
\newcolumntype{C}{>{\centering\arraybackslash}X} %
\newcolumntype{L}[1]{>{\raggedright\arraybackslash}p{#1}} %
\newcolumntype{M}[1]{>{\centering\arraybackslash}p{#1}} %

\newcommand{\boldparagraph}[1]{\vspace{0.1em}\noindent{\bf #1}}

\definecolor{darkblue}{HTML}{35394B}
\definecolor{orange}{HTML}{cc7700}
\definecolor{darkgreen}{HTML}{228B22}
\definecolor{darkgray}{HTML}{808080}
\definecolor{lightpurple}{HTML}{a56dba}
\definecolor{scratch}{HTML}{001219}
\definecolor{pretrain}{HTML}{27B9F2}

\definecolor{ours}{HTML}{E4F6FF}  %
\definecolor{c2}{HTML}{ECF8F1}
\definecolor{c1}{HTML}{DEF3E6}
\newcommand{\cellfirst}[1]{\cellcolor{c1}\textbf{#1}}
\newcommand{\cellsecond}[1]{\cellcolor{c2}{#1}}

\newcommand{\pts}{\textcolor{darkgreen}{\,$\bullet$\,}}
\newcommand{\gs}{\textcolor{pretrain}{\ding{94}\,}}

\newcommand\blfootnote[1]{%
  \begingroup
  \renewcommand\thefootnote{}\footnote{#1}%
  \addtocounter{footnote}{-1}%
  \endgroup
}

\definecolor{cvprblue}{rgb}{0.21,0.49,0.74}
\usepackage[pagebackref,breaklinks,colorlinks,allcolors=cvprblue]{hyperref}

\def\paperID{****} %
\def\confName{CVPR}
\def\confYear{2026}

\title{\includegraphics[width=0.03\textwidth, valign=m]{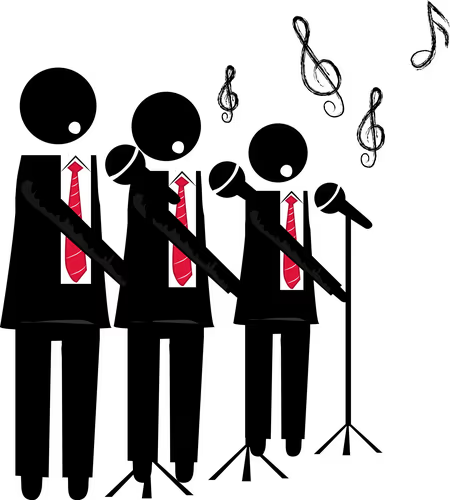} Chorus: Multi-Teacher Pretraining for Holistic 3D Gaussian Scene Encoding}

\author{  
Yue Li\textsuperscript{1,\ddag,*}, 
Qi Ma\textsuperscript{2,3,*}, 
Runyi Yang\textsuperscript{3},
Mengjiao Ma\textsuperscript{3},
Bin Ren\textsuperscript{4},
Nikola Popovic\textsuperscript{3} \\
Nicu Sebe\textsuperscript{4}, 
Theo Gevers\textsuperscript{1}, 
Luc Van Gool\textsuperscript{3},
Danda Pani Paudel\textsuperscript{3,\dag},
Martin R. Oswald\textsuperscript{1,\dag}\\
\normalsize{
\textsuperscript{1}University of Amsterdam\quad 
\textsuperscript{2}ETH Zürich\quad 
\textsuperscript{3}INSAIT, Sofia University ``St. Kliment Ohridski"} \\
\normalsize{
\textsuperscript{4}University of Trento\quad 
} \\
}

\begin{document}

\twocolumn[{%
  \renewcommand\twocolumn[1][]{#1}%
  \maketitle
  \begin{center}
  \vspace{-20pt}
  \includegraphics[width=\linewidth, trim=0 0 0pt 0, clip]{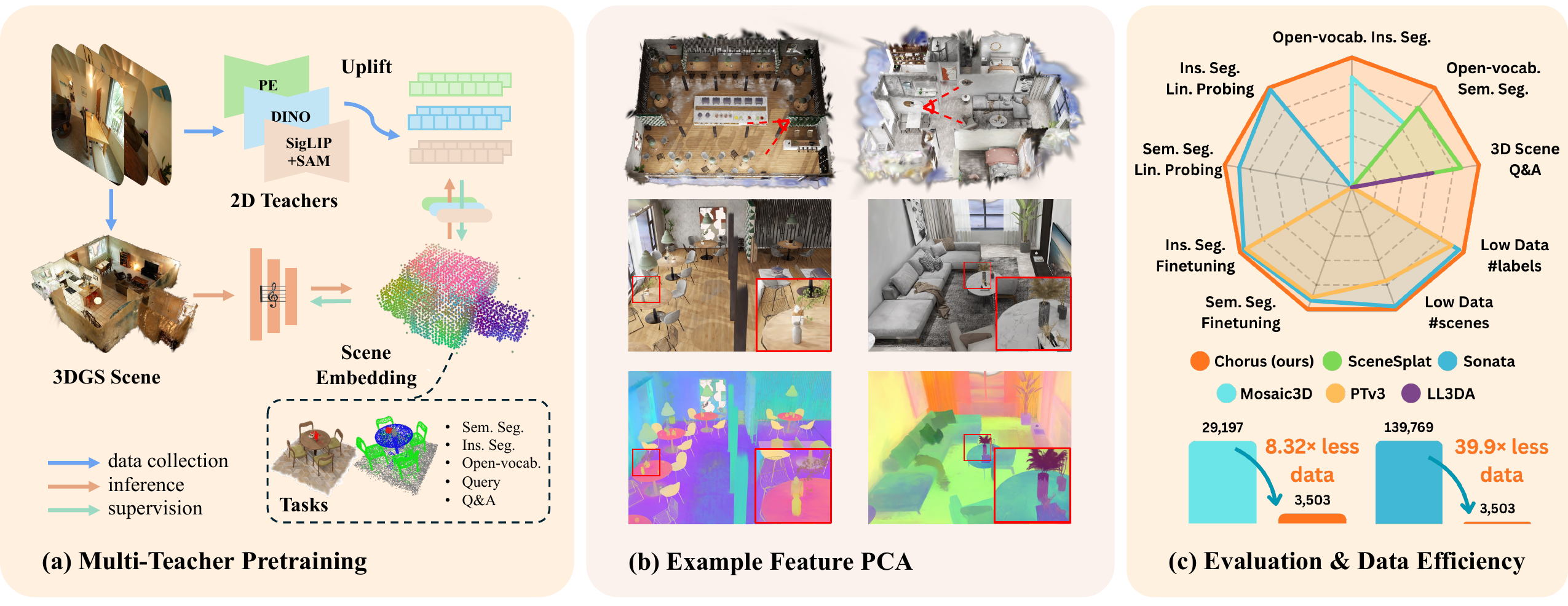}
  
  \end{center}
  \vspace{-10pt}
  \phantomsection
  \captionof{figure}{\textbf{\ours Framework.} \textbf{(a) Multi-Teacher Pretraining.} A feed-forward 3DGS scene encoder with per-teacher projectors distills complementary signals—language-aligned, generalist, and object-aware—into a shared embedding. \textbf{(b) Example Feature PCA (results on novel scenes).} At inference we input the full 3DGS scene; PCA on encoder features presents clear semantic awareness despite domain shift. \textbf{(c) Evaluation \& Data Efficiency.} \ours attains strong results across scene understanding tasks while using noticeably fewer training scenes—$8.32\times$ and $39.9\times$ less than the SoTA point-cloud pretraining baselines—highlighting the efficiency of our pretraining.}
  \vspace{1.5em}
\label{fig:teaser}
}]

{
    \blfootnote{\footnotesize \ddag{} Project lead, * Equal contribution, \dag{} Equal supervision.}
}
\addtocontents{toc}{\protect\setcounter{tocdepth}{-10}}

\begin{abstract}
While 3DGS has emerged as a high-fidelity scene representation, encoding rich, general-purpose features directly from its primitives remains under-explored. We address this gap by introducing Chorus, a multi-teacher pretraining framework that learns a holistic feed-forward 3D Gaussian Splatting (3DGS) scene encoder by distilling complementary signals from 2D foundation models.  Chorus employs a shared 3D encoder and teacher-specific projectors to learn from language-aligned, generalist, and object-aware teachers, encouraging a shared embedding space that captures signals from high-level semantics to fine-grained structure.

We evaluate \ours on a wide range of tasks: open-vocabulary semantic and instance segmentation, linear and decoder probing, data-efficient supervision, as well as LLM-based Q\&A. Besides 3DGS, we also test \ours on several benchmarks that only support point clouds by pretraining a variant using only Gaussian centers, colors, and estimated normals. Surprisingly, this encoder shows strong transfer and outperforms the point-cloud baseline while using $39.9\times$ fewer training scenes. Finally, we propose a render-and-distill adaptation that facilitates out-of-domain finetuning. Our codebase are released at this \href{https://github.com/GaussianWorld/Chorus}{repository}.
\end{abstract}

\section{Introduction}
\label{sec:intro}
The community has made rapid progress on scene representations that enable photorealistic rendering, from neural radiance fields (NeRFs)~\cite{mildenhall2020nerf} to real-time 3D Gaussian Splatting (3DGS)~\cite{kerbl20233d}. In parallel, there is a growing body of work that attaches semantic cues to these representations (\eg, by attaching vision–language features~\cite{kerr2023lerf,qin2024langsplat,marrie2025ludvig,zheng2024gaussiangrasper,peng3d}). Yet comparatively little attention has been paid to treating the 3D representation itself as a modality from which we can directly mine general-purpose, transferable features at scale. 3DGS is particularly attractive in this regard: it preserves geometry–appearance primitives and supports fast differentiable rendering, which together make it a promising substrate for large-scale pretraining beyond view synthesis~\cite{li2025scenesplat,wang2025unipre3d,jiang2025gausstr,caothousands}.

We address the gap in generalizable 3DGS scene encoding by proposing \ours{}--a multi-teacher pretraining framework for training a native 3DGS encoder to align with complementary 2D foundation models. Concretely, \ours uses a shared 3D encoder over Gaussian primitives and lightweight per-teacher projectors to distill 
\textit{(i)} \textit{language-aligned semantics} from the SigLIP2 encoder~\cite{tschannen2025siglip}, 
\textit{(ii)} \textit{generalist visual features} from DINOv3~\cite{simeoni2025dinov3}, and 
\textit{(iii)} \textit{object-aware cues} from the Perception Encoder variant PE-Spatial~\cite{bolya2025perception,kirillov2023sam}, which combines self-alignment with SAM-logit alignment to improve spatial locality while preserving semantics. 
Our multi-teacher design teaches the scene encoder breadth and complementarity, capturing high-level semantics, instance grouping, and fine spatial structure within a single 3D embedding space.

\ours builds upon the “lift-then-align” paradigm established by SceneSplat~\cite{li2025scenesplat}, which lifts dense 2D language features to 3D Gaussians and uses them as pseudo-labels to train a feed-forward 3DGS encoder for open-vocabulary segmentation. 
However, SceneSplat’s encoder was predominantly aligned with semantic information and demonstrated on semantic segmentation, leaving broader downstream applications (\eg, instance grouping) and reasoning capabilities largely unexplored.
\ours generalizes this paradigm with multi-teacher pretraining to explicitly supervise with diverse signals in order to learn a versatile 3D feature representation. Our framework therefore results in a 3D encoding that reaches superior performance across a diverse set of tasks, thereby producing \textit{a holistic 3D scene encoder}.

We demonstrate the effectiveness of \ours on the following tasks: semantic segmentation, open-vocabulary semantics, instance segmentation, open-vocabulary instances, and visual question answering. The evaluations are conducted on a comprehensive collection of datasets: ScanNet200~\cite{rozenberszki2022scannet200}, ScanNet++~\cite{scannetpp2023}, Matterport3D~\cite{chang2017matterport3d}, and our newly proposed 3DGS-native benchmark (with per-Gaussian labels) built upon InteriorGS~\cite{InteriorGS2025}.
In contrast, prior methods for generalizable 3D scene understanding typically specialize in a limited subset of tasks such as open-vocabulary tasks~\cite{lee2025mosaic3d,li2025scenesplat,ma2025scenesplat++}, semantics and instances~\cite{wu2022ptv2,wu2024ptv3,wu2025sonata}, and VQA reasoning~\cite{fu2025scene,chen2024ll3da,halacheva2025gaussianvlm}. Our evaluation demonstrates that a pretrained \ours encoder can simultaneously serve as the most effective solution across this broad spectrum of scene understanding tasks.
Furthermore, we carry out probing experiments (linear/decoder probing and full finetuning) for semantic and instance segmentation to assess the feature quality across the same datasets.
In addition, we conduct data-efficiency studies that restrict supervision to limited scenes and sparse annotations, thereby stress-testing how much the pretrained encoder alone carries.

Besides 3DGS, we tested \ours on several benchmarks that only support point clouds. For this purpose, we pretrained a new point-cloud-compatible 3D encoder using the Gaussians’ centers, colors, and estimated normals as the only inputs, while keeping all other training signals and losses identical. To our surprise, this variant is competitive with the recent self-supervised point cloud pretraining method Sonata~\cite{wu2025sonata} while using $\sim39.9\times$ fewer training scenes. \ours also exhibits favorable scaling as we move from subset to joint-dataset pretraining. These observations indicate that our multi-teacher pretraining successfully mines semantics, spatial locality, and instance grouping that carry over when the encoder is evaluated on point-cloud tasks, despite the distribution difference. In practice, multi-teacher distillation over 3DGS is a practical, efficient route towards a general feed-forward scene encoder.

To further demonstrate versatility, \ours facilitates out-of-domain adaptation by introducing a render-and-distill strategy that eliminates the need for costly 3D pseudo-label preprocessing. This approach leverages the inherent rendering capability of 3DGS: given a new dataset, we simply render 2D views, perform online teacher inference, and finetune our encoder with teacher knowledge. This makes the adaptation pipeline more lightweight and accessible.

Our contributions can be summarized as follows: 

\begin{itemize}
    \item  A multi-teacher pretraining framework that aligns a native 3DGS encoder with diverse 2D teachers (language-aligned, generalist, and object-aware) via a shared backbone and per-teacher projectors.

    \item A holistic 3D scene encoding that produces highly structured and transferable embeddings for both 3DGS and PC inputs, leading to state-of-the-art performance on a broad range of tasks, while demonstrating data efficiency.

    \item A lightweight render-and-distill adaptation recipe that enables convenient out-of-domain finetuning without requiring costly 3D pseudo-label preprocessing.
\end{itemize}

\section{Related Work}
\label{sec:relatedwork}

\begin{figure*}
    \centering
    \includegraphics[width=\linewidth]{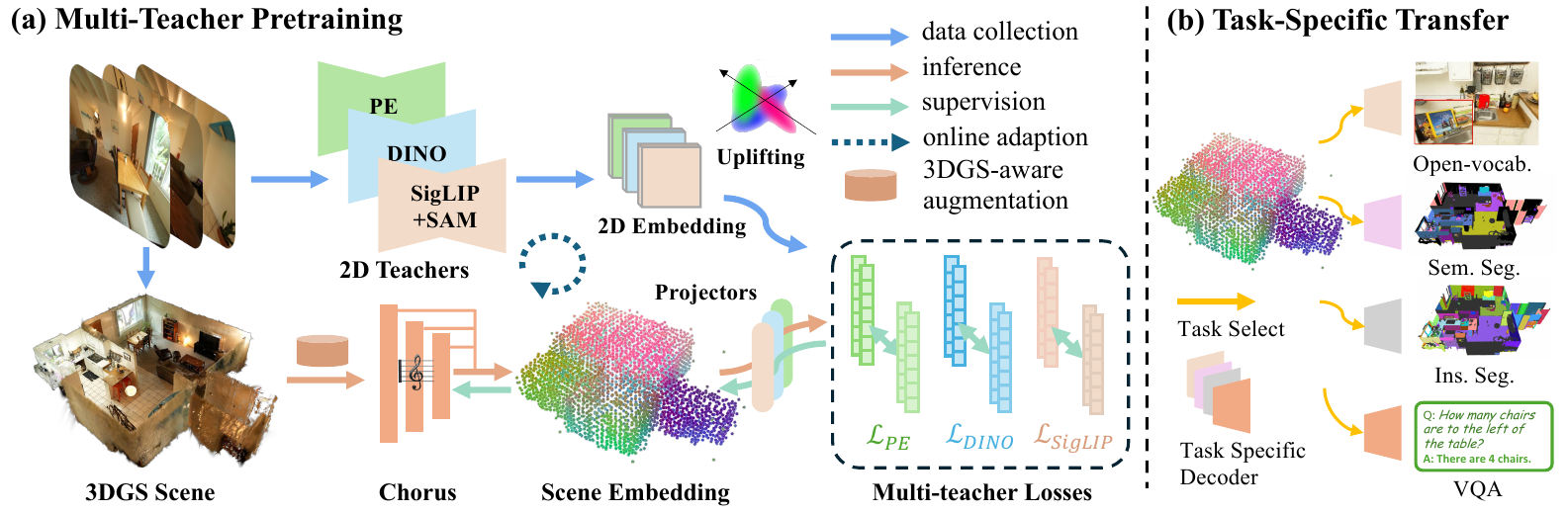}
    \caption{\textbf{\ours Overview.} \textbf{(a) Multi-Teacher Pretraining.} We train a feed-forward 3DGS scene encoder to distill complementary signals--language-aligned (SigLIP), generalist (DINO), and object-aware (PE)--from 2D teachers. This knowledge is transferred into a shared embedding space via lightweight per-teacher projectors and losses. To accelerate out-of-domain adaptation, we support finetuning the encoder with online rendering-based supervision. \textbf{(b) Task-Specific Transfer.} A pretrained \ours encoder enables diverse downstream tasks, including semantic and instance segmentation, open-vocabulary query, and 3D visual question answering (VQA).}
    \label{fig:method_Diagram}
    \vspace{-10pt}
\end{figure*}

\boldparagraph{Self-supervised and Cross-modal Distillation for 3D.}
Self-supervised learning (SSL) has driven strong representation learning for 2D images~\cite{he2020momentum,he2022masked,caron2021emerging} and has been actively explored for 3D data via contrastive and masked modeling~\cite{xie2020pointcontrast,yu2022point,afham2022crosspoint,pang2023masked,ren2024bringing}. Recently, Sonata~\cite{wu2025sonata} mitigates the geometric shortcut during point-cloud self-supervised learning, and~\cite{zhang2025concerto} shows that joint 2D–3D SSL can yield more coherent spatial features than using a single modality.
In parallel, knowledge distillation~\cite{hinton2015distilling} has emerged as a powerful paradigm. Cross-modal distillation injects priors from 2D foundation models~\cite{radford2021learning,oquab2023dinov2,kirillov2023segment,zhai2023siglip,tschannen2025siglip} into 3D, mitigating label scarcity and enabling semantic awareness in 3D representations~\cite{peng2023openscene,xue2023ulip,kerr2023lerf,xu2024pointllm}.
This distillation paradigm has progressed from single-teacher to multi-teacher aggregation in the 2D domain, as shown by~\cite{ranzinger2024radio,heinrich2025radiov2,sariyildiz2025dune}, which strengthens learning with complementary signals.
We adopt this perspective and specialize it to 3DGS for the first time: instead of a single teacher or objective, \ours distills from various 2D teachers (language-aligned, generalist, object-aware) to align embeddings with rich priors, while leveraging the inherent rendering capability of 3DGS.

\boldparagraph{3D Gaussian Splat Encoders.}
Unlike 3D point clouds, where representation learning is well explored~\cite{qi2017pointnet,qi2017pointnet++,zhao2021point,wu2022ptv2,wu2024ptv3}, encoding 3D Gaussian Splats remains under-explored despite their richer parameter space that couples both appearance and geometry. ShapeSplat~\cite{ma2025large} pioneers object-level masked reconstruction for 3DGS objects, while Can3Tok~\cite{gao2025can3tok} learns a scene-level VAE that tokenizes 3DGS scenes into latent codes. At the scene level, SceneSplat~\cite{li2025scenesplat} lifts 2D semantic priors to train a generalizable 3DGS encoder for open-vocabulary semantics and introduces the SceneSplat-7K dataset. SceneSplat++~\cite{ma2025scenesplat++} further scales scene-level 3DGS data and establishes a comprehensive benchmark for language-aligned 3DGS methods. Building on this trajectory, \ours proposes consolidating complementary 2D priors into a single feed-forward 3DGS encoder, producing holistic scene embeddings that transfer robustly across diverse tasks (semantics, instances, and question answering~\cite{halacheva2025gaussianvlm}).

\section{Method}
\label{sec:method}
\ours pretrains a general-purpose feed-forward Gaussian scene encoder by distilling knowledge from multiple 2D teachers. We first explain the pretraining data, where the 2D feature maps are lifted to the 3DGS (\S\ref{subsec:lifting}). Then we present the multi-teacher framework, \ie, a shared 3DGS encoder with lightweight per-teacher projectors and losses, including optional contrastive terms that exploit available semantic class/instance structure (\S\ref{subsec:multi_teacher}). Next, we describe a rendering-based adaptation recipe that shortcuts adaptation via image-plane supervision, accelerating out-of-distribution generalization (\S\ref{subsec:render_adapt}). Finally, we introduce our 3DGS-aware augmentations to aid pretraining (\S\ref{subsec:3dgs_aug}).

\subsection{Lifting 2D Teachers for Supervision}
\label{subsec:lifting}
\boldparagraph{3DGS scene rendering.}
A 3DGS scene is an optimized parameter set of $N$ Gaussians:
\begin{equation}
\textstyle
\mathcal{G}=\{(\mathbf{x}_i,\mathbf{s}_i,\mathbf{q}_i,\alpha_i,\mathbf{c}_i)\}_{i=1}^N
\end{equation}
to reproduce multi-view images via alpha composition and
anisotropic Gaussians~\cite{kerbl20233d}.
Each tuple contains parameters for a center $\mathbf{x}_i\!\in\!\mathbb{R}^3$, scale $\mathbf{s}_i\!\in\!\mathbb{R}^3_{+}$, orientation $\mathbf{q}_i\!\in\!\mathbb{H}$ (unit quaternion), opacity $\alpha_i\!\in\![0,1]$, and color $\mathbf{c}_i\!\in\![0,1]^3$.
For a viewpoint $p$ and a pixel $\mathbf{u}\in\mathbb{N}^2$, 3DGS renders colors as
\begin{equation}
\label{eq:3dgs_render}
\mathbf{C}(\mathbf{u}\!\mid\! p)\!=\!\sum\limits_{i\in\mathcal{S}_{d,\mathbf{u}}}
\underbrace{T_i\,\alpha_i(\mathbf{u}\!\mid\! p)}_{w_i(p,\mathbf{u})}\;\mathbf{c}_i,\;\;
T_i=\prod\limits_{j<i}\big(1-\alpha_j(\mathbf{u}\!\mid\! p)\big),
\end{equation}
where $\mathcal{S}_{d,\mathbf{u}}$ is the depth-sorted set of splats intersecting the viewing ray.

\boldparagraph{Normalized uplifting.}
Let $F_{p,\mathbf{u}}$ be a 2D teacher feature at view $p$, pixel $\mathbf{u}$, and let $f_i$ be the target feature on Gaussian $i$.
Using the same rendering weights as in~\eqref{eq:3dgs_render}, we obtain uplifted supervision as a weighted average~\cite{marrie2025ludvig}:
\begin{equation}
\label{eq:uplift}
\textstyle
f_i=\!\!\sum\limits_{(p,\mathbf{u})\in\mathcal{S}_i}\!\!\!\!\!\bar{w}_i(p,\mathbf{u})\,F_{p,\mathbf{u}},
\;\;
\bar{w}_i(p,\mathbf{u})=\frac{w_i(p,\mathbf{u})}{\!\!\!\!\!\sum\limits_{(p',\mathbf{u}')\in\mathcal{S}_i} \!\!\!\!\!\! w_i(p',\mathbf{u}')},
\end{equation}
where $\mathcal{S}_i$ is the set of all view-pixel pairs contributing to feature $f_i$.

\boldparagraph{Teacher standardization.}
We supervise with three 2D teachers: SigLIP2 (\emph{language-aligned}), DINOv3 (\emph{generalist features}), and PE-Spatial (\emph{object-aware}). Because teacher activations differ in scale/variance, we apply PHI-S~\cite{ranzinger2024phi}, a PCA rotation followed by isotropic Hadamard scaling to achieve unit average per-channel variance while preserving cross-channel relationships. We denote $\,\widetilde{F}_{p,\mathbf{u}}=\text{PHI\mbox{-}S}(F_{p,\mathbf{u}})\,$ and use $\widetilde{f}_i$ analogously when needed.

\subsection{Multi-Teacher Pretraining}
\label{subsec:multi_teacher}
\boldparagraph{Architecture.}
A shared 3DGS encoder $g_\theta$ maps Gaussian parameters to latent per-Gaussian features:
\begin{equation}
\textstyle
Z = g_\theta(\mathcal{G}) \in \mathbb{R}^{N\times d_z}.
\end{equation}
Each teacher $t\in\mathcal{T}=\{\text{lang},\text{dino},\text{pe}\}$ has a lightweight projector head $h_t$ (2-layer MLP with $\mathrm{LayerNorm}$ and $\mathrm{GELU}$) producing per-Gaussian predictions
$
\hat{f}^{(t)}\!=\!h_t(Z)\in\mathbb{R}^{N\times d_t}.
$

\boldparagraph{Per-teacher losses.}
For teacher $t$, we denote the uplifted per-Gaussian supervision $\widetilde{f}^{(t)}$ and an optional validity mask $M^{(t)}$ (derived from feature norms and visibility). Our base matching loss combines cosine and smooth-$\ell_1$ loss:
\begin{align}
\label{eq:match}
\mathcal{L}_{\text{match}}^{(t)} &=
\frac{1}{|M^{(t)}|}
\sum_{i\in M^{(t)}}
\lambda_1\Big(1-\cos\big(\hat{f}^{(t)}_i,\widetilde{f}^{(t)}_i\big)\Big) \notag\\
&\quad + \lambda_2\, \mathrm{SmoothL1}\big(\hat{f}^{(t)}_i,\widetilde{f}^{(t)}_i\big),
\end{align}
to preserve both magnitude and angular alignment. We $\ell_2$-normalize the inputs before calculating cosine terms.

\boldparagraph{Teacher-specific contrastive loss (optional).}
We add compact contrastive regularizers~\cite{li2025scenesplat} when the source dataset provides semantic/instance labels.
\begin{itemize}
\item SigLIP2 teacher (semantic): pool class-wise means
$
\bar{f}^{(t)}_c {=} \text{mean}\{\hat{f}^{(t)}_i:i\!\in\!\mathcal{G}_c\}
$,
split each class into two disjoint halves $A/B$, and apply a bidirectional InfoNCE loss over $\ell_2$-normalized $\bar{f}^{(t)}_{c,\{A,B\}}$ across classes.
\item PE-Spatial teacher (instance): pool instance-wise means
$
\bar{f}^{(t)}_k {=} \text{mean}\{\hat{f}^{(t)}_i:i\!\in\!\mathcal{I}_k\}
$
and similarly apply InfoNCE.
\end{itemize}
We write the loss term succinctly as $\mathcal{L}_{\text{con}}^{(t)}$ and provide the equations in the supplement. 

\boldparagraph{Staged pretraining \& total optimization objective.}
Teachers can start at different epochs. Let $\mathcal{A}(e)\!\subseteq\!\mathcal{T}$ denote the active set at training epoch $e$ (\eg, $\{\text{lang},\text{dino}\}$ from the start, then add $\text{pe}$). The total objective is
\begin{equation}
\label{eq:total}
\textstyle
\mathcal{L}_{\text{total}}(e)=
\sum_{t\in\mathcal{A}(e)} \lambda_t
\Big(\mathcal{L}_{\text{match}}^{(t)} + \eta_t\,\mathcal{L}_{\text{con}}^{(t)}\Big),
\end{equation}
with a simple per-teacher weight $\lambda_t$ and optional $\eta_t$. We empirically found that PHI-S standardization simplifies loss balancing, \ie, equal weights of $\lambda_t$ suffice across teachers.

\subsection{Rendering-Based Adaptation}
\label{subsec:render_adapt}
Given a novel data domain, we can adapt our pretrained encoder without precomputing 3D pseudo-labels via online inference. We sample camera poses $\{p\}$, conduct visibility culling on the input Gaussians, run each 2D teacher on the rendered RGB to obtain feature maps $F^{(t)}_{p,\mathbf{u}}$, and obtain per-Gaussian predictions $\hat{f}^{(t)}_i$ with the current encoder and projector heads.
Using the same compositing weights $w_i(p,\mathbf{u})$ as in Eq.~\eqref{eq:3dgs_render}, we render an inference feature map for each teacher $t$:
\begin{equation}
\label{eq:adapt_render}
\textstyle
\hat{F}^{(t)}_{p,\mathbf{u}}
=\sum_{i\in\mathcal{S}_{p,\mathbf{u}}} w_i(p,\mathbf{u})\,\hat{f}^{(t)}_{i}.
\end{equation}

\boldparagraph{Adaptation objective.}
Let $\Omega$ be the set of valid pixels with sufficient transmittance. We reuse the same per-teacher matching loss as in Eq.~\eqref{eq:match} (cosine + SmoothL1, with the same $\lambda_1,\lambda_2$), now applied to the 2D feature maps over $\Omega$:
\begin{equation}
\label{eq:adapt_loss}
\textstyle
\mathcal{L}^{(t)}_{\text{img}}
=\frac{1}{|\Omega|}
\sum_{(p,\mathbf{u})\in\Omega}
\ell_{\text{match}}\!\big(\hat{F}^{(t)}_{p,\mathbf{u}},\,\widetilde{F}^{(t)}_{p,\mathbf{u}}\big).
\end{equation}
This \emph{render-and-distill} loop adapts \ours using only rendered frames, accelerating the adaptation to new data.

\begin{figure}
    \centering
    \includegraphics[width=\linewidth]{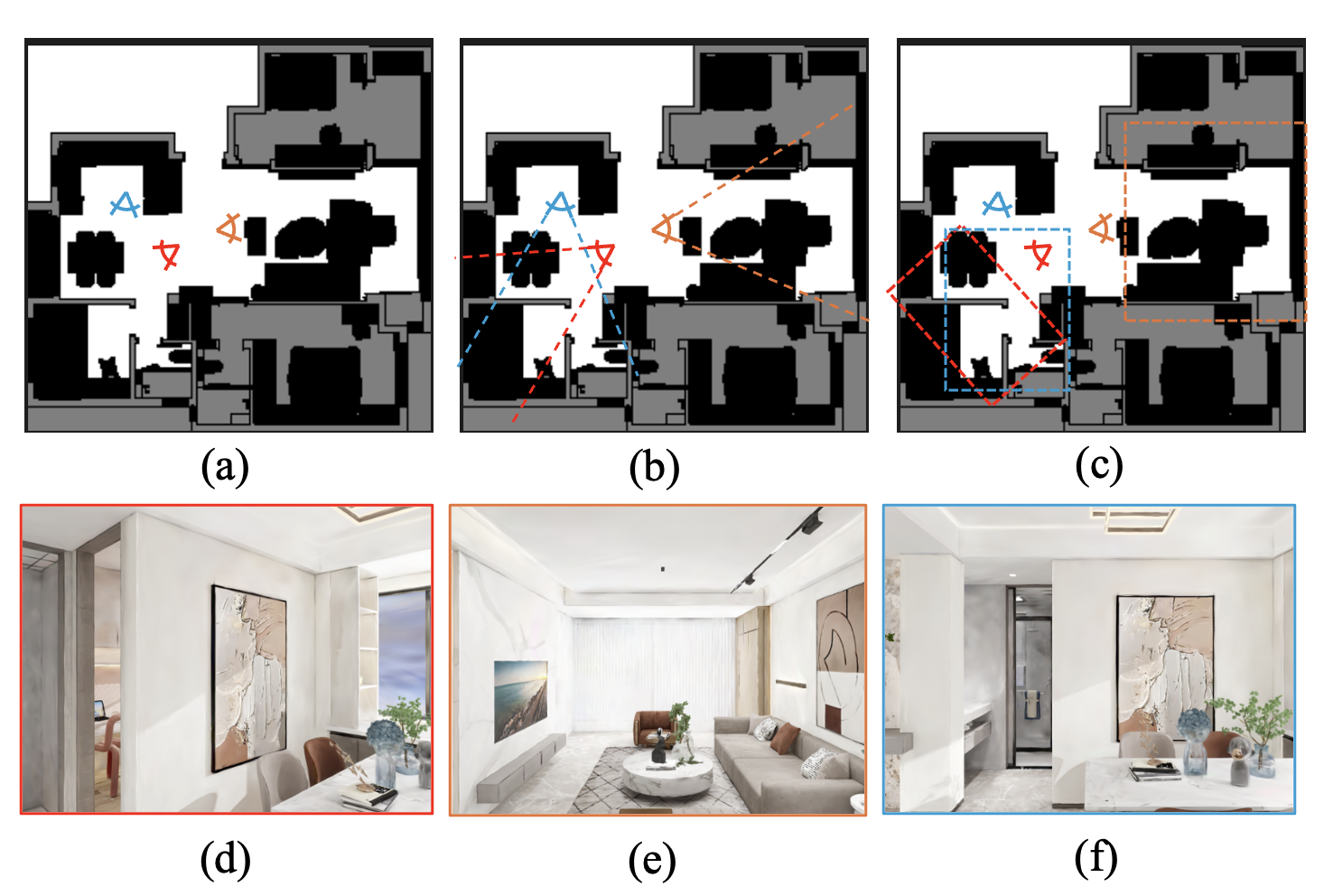}
    \caption{\textbf{Rendering-Based View Sampling and Pairing:} (a) Camera Location Sampling: We use Furthest Point Sampling to select camera positions that achieve broad spatial coverage across the entire \textit{navigable} scene space. (b) Visibility Culling: For each location, we sample view angles and track the \textit{visibility} of the 3D Gaussians across frames. (c) View Pairing and Selection: We obtain a minimum \textit{2D bounding box} covering all visible Gaussians for a given view. Candidate pairs of poses are then calculated and sorted based on the overlap score. (d,e,f) Rendered images corresponding to the colored camera viewpoints. }
    \label{fig:2D_supervise}
\end{figure}

\boldparagraph{View sampling and pairing pipeline.}
To enable adaptation on data without provided poses, we select informative views for 2D supervision in two stages: (1) \textit{Informative view selection} -- sample camera positions that are well distributed in the navigable space yet are neither too close to geometry nor heavily occluded; (2) \textit{Contextual view pairing} -- ensure that selected views share sufficient overlap to promote cross-view feature coherence. 
As illustrated in \cref{fig:2D_supervise}, we first sample camera positions proportional to the scene size to ensure coverage, and for each position generate eight candidate horizontal viewing directions. Directions whose center ray is too close to scene contents are discarded. For each valid view, we rasterize the Gaussians and record their visibility, then compute the minimum 2D axis-aligned bounding box enclosing all visible splats; only Gaussians that fall inside this region are kept as input for that training view. Finally, for each camera pose we sort the remaining poses by visibility overlap and form training pairs from high-overlap poses, ensuring sufficient multi-view context. Further details are provided in the supplement.

\subsection{3DGS-Aware Augmentations}
\label{subsec:3dgs_aug}
\boldparagraph{Why are point-cloud augmentations suboptimal for 3DGS?}
Point-cloud augmentations (dropout, elastic distortions, color/geometry jitter, etc.) are designed for i.i.d.\ sets of points whose attributes (position, color) are direct observations of 3D geometry/appearance. In contrast, a 3D Gaussian Splatting (3DGS) scene is an \emph{optimized} parameter space. Naïve point-cloud jitter alters $\alpha_i$ and $T_i$ in ways that are not motivated for splat-based rendering, and empirically we observe consistent performance drops when such jitter is applied to our encoder pretraining.

\boldparagraph{Design principle.}
We propose two augmentations that are \emph{3DGS-aware}: (i) a \emph{Rendering-Equivalent} perturbation, which targets augmented parameters that render approximately the same images and injects small, covariance-aware position noise primarily into low-opacity splats; and (ii) an \emph{Immature-Manifold} perturbation, which mimics earlier (blurrier) stages of optimization by selectively inflating per-splat covariances. Both augmentations are grounded in the rendering equation~\cref{eq:3dgs_render} and the observed 3DGS optimization dynamics~\cite{kheradmand20243d,jung2024relaxing}; equations are provided in the supplement.

\section{Experiments}
\label{sec:experiments}
We evaluate the pretrained \ours encoder on diverse tasks. First, using the language-aligned projector, we report open-vocabulary semantic and instance segmentation. Next, we show that \ours serves as an effective scene tokenizer for the language model, enabling question answering (\S\ref{subsec:main_results}). 
Later, we pretrain a point-cloud variant of \ours using only Gaussian centers, colors, and estimated normals as input; it surprisingly competes with SoTA point-cloud encoders (\S\ref{subsec:pts_results}).
We analyze this unexpected robustness in our ablation study (\S\ref{subsec:ablation}).

\boldparagraph{Implementation details.}
Our pretraining backbone adapts the 5-stage transformer encoder from Sonata~\cite{wu2025sonata} with a 512-dimensional bottleneck feature. We employ the teachers \texttt{SigLIP2-so400m-p16-512}, \texttt{DINOv3-ViTL16}, and \texttt{PE-Spatial-L14-448}. For pretraining data, we use collected 3DGS scenes (center, color, opacity, quaternion, scale) from SceneSplat-7K~\cite{li2025scenesplat}, with newly processed pseudo-labels for each teacher. We set teacher loss weights $\lambda_t = 1.0$ and balance the contrastive terms with $\eta_t = 0.02$ for $\mathcal{L}_{\text{con}}^{(t)}$. We pretrain the standard model (denoted \gs) using all 3DGS parameters as input, and a point-cloud variant (denoted \pts) using only Gaussian centers, colors, and estimated normals; all other settings are unchanged. This variant is used for subsequent probing and finetuning to compare against point-cloud encoders. %
For rendering-based adaptation, we initialize with the pretrained \ours encoder and select 4 overlapping views per batch. By default, the rendered image resolution is $480\times640$, and the rendered feature resolution is $120\times160$. We bilinearly upsample the online-encoded 2D teacher features to match the rendered feature map. The adaptation uses a learning rate of $2 \times 10^{-4}$ and runs for 100 epochs. The SceneSplat-7K MP3D~\cite{chang2017matterport3d} subset was updated for \ours training. Consequently, we re-trained \cite{li2025scenesplat} using public code for joint-data comparison.
We refer to the supplement for additional training and experimental details.

\subsection{Main Results}

\label{subsec:main_results}

\begin{table*}[t]
\centering

\begingroup
\setlength{\tabcolsep}{6pt}
\footnotesize                      %

\begin{tabular}{ll|r|rr|rr|rr|rr}
\toprule[0.95pt]
\multirow{2}{*}{Method}
& \multirow{2}{*}{Training Source}
& \multirow{2}{*}{\shortstack{\#Training \\ Scenes}}
& \multicolumn{2}{c|}{ ScanNet200 (200)}
& \multicolumn{2}{c|}{ Matterport3D (160)}
& \multicolumn{2}{c|}{ ScanNet++ (100)}
& \multicolumn{2}{c}{\gs InteriorGS (72)}
\\
& & & f-mIoU & f-mAcc & f-mIoU & f-mAcc & f-mIoU & f-mAcc & f-mIoU & f-mAcc \\
\midrule[0.6pt]
OpenScene$^{\dagger}$~\cite{peng2023openscene}
& SN & $\times$1
& 6.4 & 12.2
& 5.7 & 10.7
& 8.8 & 14.7
& -- & -- \\
PLA~\cite{ding2023pla}
& SN & --
& 1.8 & 3.1
& -- & --
& -- & --
& -- & -- \\
RegionPLC~\cite{yang2024regionplc}
& SN & --
& 9.2 & 16.4
& 6.2 & 13.3
& 11.3 & 20.1
& -- & -- \\
OV3D~\cite{jiang2024open}
& SN & --
& 8.7 & --
& -- & --
& -- & --
& -- & -- \\
Mosaic3D~\cite{lee2025mosaic3d}
& SN & --
& {13.0} & {24.5}
& {8.6}  & {17.8}
& {16.2} & {27.1}
& {3.8}  & {8.2} \\
\gs SceneSplat~\cite{li2025scenesplat}
& SN & --
& {18.9} & {31.7}
& {10.8} & {18.7}
& {14.7} & {24.7}
& {6.1}  & {8.5} \\
{\cellcolor{ours}}\gs \ours (ours)
& SN & --
& {22.4} & {45.8}
& {11.4} & {16.4}
& {16.8} & {29.1}
& {9.0}  & {14.6} \\
\midrule[0.6pt]
\multirow{2}{*}{Mosaic3D~\cite{lee2025mosaic3d}}
& \multirow{2}{*}{\makecell[l]{SN, SN++, MP3D,\\  ARKitS, S3D}}
& \multirow{2}{*}{$\times$24.3}
& \multirow{2}{*}{{15.7}} & \multirow{2}{*}{{28.3}}
& \multirow{2}{*}{{13.1}} & \multirow{2}{*}{{27.7}}
& \multirow{2}{*}{18.0}   & \multirow{2}{*}{29.0}
& \multirow{2}{*}{9.4}    & \multirow{2}{*}{16.0} \\
& & & & & & & & & & \\
\gs SceneSplat~\cite{li2025scenesplat}
& SN, SN++, MP3D & $\times$2.92
& \cellsecond{22.5} & \cellsecond{41.7}
& \cellsecond{14.0} & \cellsecond{32.4}
& \cellsecond{28.6} & \cellsecond{50.9}
& \cellsecond{10.0} & \cellsecond{18.3} \\
{\cellcolor{ours}}\gs \ours (ours)
& SN, SN++, MP3D & $\times$2.92
& \cellfirst{24.6} & \cellfirst{47.7}
& \cellfirst{18.7} & \cellfirst{38.5}
& \cellfirst{29.6} & \cellfirst{53.5}
& \cellfirst{15.7} & \cellfirst{24.1} \\
\bottomrule[0.95pt]
\end{tabular}
\endgroup

\caption{
    \textbf{Zero-Shot 3D Semantic Segmentation on the Fine-Grained ScanNet++ (100 classes)~\cite{scannetpp2023}, Matterport3D (160 classes)~\cite{chang2017matterport3d}, ScanNet200 (200 classes)~\cite{dai2017scannet}, and InteriorGS (72 classes)~\cite{InteriorGS2025} Benchmarks.}
    \gs denotes 3DGS modality input. %
    \ours and SceneSplat~\cite{li2025scenesplat} are the \emph{only} methods that target 3DGS modality pretraining.
    We report the foreground mean IoU (f-mIoU) and foreground mean accuracy (f-mAcc) excluding the wall, floor, and ceiling classes, following~\cite{yang2024regionplc,peng2023openscene}.
    $^{\dagger}$ denotes the official checkpoint and the baseline results are partly taken from~\cite{lee2025mosaic3d}.
    Dataset abbreviations SN, SN++, ARKitS, MP3D, and S3D are short for ScanNet~\cite{dai2017scannet}, ScanNet++~\cite{scannetpp2023},
    ARKitScenes~\cite{baruch2021arkitscenes}, Matterport3D~\cite{chang2017matterport3d}, and Structured3D~\cite{zheng2020structured3d}. 
    \ours achieves noticeably better zero-shot performance, \eg, a $2.1\%$ f-mIoU and $6.0\%$ f-mAcc increase on ScanNet200, and when evaluated on new data, a $5.7\%$ f-mIoU and $5.8\%$ f-mAcc increase on \interior compared to the previous SoTA SceneSplat, while using $8.32\times$ less training data than the point-cloud-based pretraining method~\cite{lee2025mosaic3d}.}
    \vspace{-2mm}
\label{tab:semseg_zeroshot}
\end{table*}

\begin{table}[ht]
\footnotesize
\centering
\setlength{\tabcolsep}{2pt}
\renewcommand{\arraystretch}{1.05}

\begin{tabularx}{\linewidth}{l l l *{4}{C}}
\toprule
\renewcommand\cellgape{} %

\multirow{2}{*}{Method} &
\multirow{2}{*}{Inputs} &
\multirow{2}{*}{\makecell[l]{3D Region Proposal\\Network}} &
$\text{mAP}$ &
$\text{mAP}$ &
$\text{mAP}$ &
$\text{mAP}$
\\
& & & \scriptsize$\text{25}$ & \scriptsize$\text{50}$ & \scriptsize$\text{head}$ & \scriptsize$\text{tail}$ \\
\midrule
Open3DIS & 3D+2D & {\makecell[l]{Superpoints + ISBNet\\+ Grounded-SAM}} & 32.8 & 29.4 & 27.8 & 21.8 \\
SAI3D & 3D+2D & Superpoints + SAM & 24.1 & 18.8 & 12.1 & 16.2 \\
\midrule
OpenScene-3D & 3D & Mask3D & 7.2 & 6.2 & 10.6 & 0.7 \\
RegionPLC     & 3D & Mask3D & 9.7 & 8.6 & 15.6 & 1.7 \\
OpenIns3D     & 3D & Mask3D & 14.4 & 10.3 & 16.0 & 4.2 \\
Mosaic3D      & 3D & Mask3D & \cellsecond{17.8} & \cellsecond{16.0} & \cellfirst{21.8} & \cellsecond{5.4} \\
{\cellcolor{ours}}\gs \ours (ours) & 3D & Mask3D & \cellfirst{19.6} & \cellfirst{18.0} & \cellsecond{18.5} & \cellfirst{13.0} \\
\bottomrule
\end{tabularx}

\caption{\textbf{Open-Vocabulary 3D Instance Segmentation on ScanNet200.} Methods are grouped by input type. Methods using both 3D+2D inputs require expensive multi-view image processing, whereas \ours is feed-forward and shows strong performance, especially on the 66 tail classes.}
\label{tab:insseg_open_vocab}
\end{table}

\begin{table}[t]
\centering
\begingroup
\scriptsize
\renewcommand{\arraystretch}{1.05}

\begin{tabularx}{\linewidth}{L{25mm} *{6}{C}}
\toprule
\multirow{2}{*}[-0.7ex]{Methods} & \multicolumn{3}{c}{\textbf{ScanQA}} & \multicolumn{3}{c}{\textbf{Nr3D}} \\
\cmidrule(lr){2-4} \cmidrule(lr){5-7}
& EM1 & M & \multicolumn{1}{c}{R} & Sim & M & R \\

\midrule
ScanQA~\cite{azuma2022scanqa}               & --   & 13.1 & 33.3 & --   & --   & --   \\
3D-VLP~\cite{yang20243d}                    & --   & 13.5 & 34.5 & --   & --   & --   \\
Scene-LLM~\cite{fu2025scene}                & --   & 15.8 & --   & --   & --   & --   \\
LL3DA~\cite{chen2024ll3da}                 & 14.3 & 22.8 & 34.7 & 48.1 & 5.8  & 9.9  \\
\makecell[l]{GaussianVLM~\cite{halacheva2025gaussianvlm}\\$\vdash$ \gs SceneSplat*~\cite{li2025scenesplat}}
                                            & 14.4 & \cellfirst{22.9} & 34.8 & 48.2 & 20.8 & 19.2 \\
{\cellcolor{ours}\makecell[l]{GaussianVLM\\$\vdash$ \gs \ours\ (ours)}}
                                            & \cellfirst{14.8} & 22.5 & \cellfirst{37.4} & \cellfirst{50.6} & \cellfirst{22.5} & \cellfirst{28.8} \\
\bottomrule
\end{tabularx}
\endgroup

\caption{\textbf{3D Scene Question Answering.} Comparison across ScanQA (EM@1/M/R) and Nr3D (Sim/M/R).} %
\label{tab:scan_all}
\end{table}

\begin{table*}[t]
\centering
\footnotesize
    \begin{minipage}{\textwidth}
    \centering
        \begingroup
\setlength{\tabcolsep}{4pt} 
\footnotesize                      

\begin{tabularx}{\textwidth}{l *{5}{CCC}}
\toprule
Probing Exp.
& \multicolumn{3}{c}{ ScanNet Val}
& \multicolumn{3}{c}{ ScanNet200 Val}
& \multicolumn{3}{c}{ ScanNet++ Val}
& \multicolumn{3}{c}{ Matterport3D (160)}
& \multicolumn{3}{c}{\gs InteriorGS} \\
\cmidrule(lr){1-1}
\cmidrule(lr){2-4}
\cmidrule(lr){5-7}
\cmidrule(lr){8-10}
\cmidrule(lr){11-13}
\cmidrule(lr){14-16}
Methods
& mIoU & mAcc & allAcc
& mIoU & mAcc & allAcc
& mIoU & mAcc & allAcc
& mIoU & mAcc & allAcc
& mIoU & mAcc & allAcc \\
\midrule

MSC~\cite{wu2023msc} (lin.)
& 21.8 & 32.2 & 65.5
& 3.3  & 5.5  & 57.5
& 8.1  & 11.9 & 64.7
& -- & -- & --
& -- & -- & -- \\

Sonata~\cite{wu2025sonata} (lin.)
& 73.7 & 84.4 & 90.3
& 28.8 & 38.8 & 81.8
& 40.7 & 55.3 & 84.8
& 18.5 & \cellfirst{25.8} & 78.8
& 24.3 & 35.4 & 61.4 \\

\cellcolor{ours}\gs/\pts \ours (lin.)
& \cellfirst{75.2} & \cellfirst{84.8} & \cellfirst{90.5}
& \cellfirst{36.0} & \cellfirst{47.2} & \cellfirst{82.8}
& \cellfirst{48.8} & \cellfirst{63.2} & \cellfirst{86.4}
& \cellfirst{20.0} & 25.7 & \cellfirst{79.4}
& \cellfirst{27.0} & \cellfirst{37.2} & \cellfirst{62.6} \\

\cmidrule(lr){1-16}

Sonata~\cite{wu2025sonata} (dec.)
& \cellfirst{77.3} & \cellfirst{85.9} & \cellfirst{92.0}
& 30.1 & 39.4 & \cellfirst{83.0}
& 46.6 & 58.9 & \cellfirst{86.8}
& 19.0 & 26.2 & 79.4
& 27.2 & 38.6 & 65.3 \\

\cellcolor{ours}\gs/\pts \ours (dec.)
& 75.0 & 83.1 & 90.6
& \cellfirst{32.5} & \cellfirst{43.0} & 82.2
& \cellfirst{48.4} & \cellfirst{62.3} & 86.7
& \cellfirst{19.6} & \cellfirst{26.4} & \cellfirst{79.6}
& \cellfirst{29.3} & \cellfirst{41.6} & \cellfirst{66.8} \\

\midrule

\textcolor{darkgray}{PTv3 (sup)}~\cite{wu2024ptv3}
& \textcolor{darkgray}{77.4} & \textcolor{darkgray}{84.8} & \textcolor{darkgray}{92.0}
& \textcolor{darkgray}{34.7} & \textcolor{darkgray}{45.4} & \textcolor{darkgray}{83.5}
& \textcolor{darkgray}{48.2} & \textcolor{darkgray}{61.6} & \textcolor{darkgray}{87.0}
& \textcolor{darkgray}{17.5} & \textcolor{darkgray}{23.3} & \textcolor{darkgray}{78.9}
& \textcolor{darkgray}{31.1} & \textcolor{darkgray}{44.0} & \textcolor{darkgray}{67.4} \\
MSC (f.t.)~\cite{wu2023msc}
& 78.2 & 85.3 & 92.2
& 33.4 & 43.7 & 83.4
& 48.7 & 61.9 & 87.2
& -- & -- & --
& -- & -- & -- \\
Sonata (f.t.)~\cite{wu2025sonata}
& 78.6 & 86.6 & 92.3
& 34.4 & 44.0 & 84.0
& 49.9 & 60.7 & \cellfirst{87.4}
& 21.3 & 27.6 & 80.2
& 30.7 & 41.8 & 66.2 \\
\cellcolor{ours}\gs/\pts \ours (f.t.)
& \cellfirst{79.4} & \cellfirst{87.7} & \cellfirst{92.4}
& \cellfirst{40.9} & \cellfirst{52.3} & \cellfirst{84.1}
& \cellfirst{52.9} & \cellfirst{66.2} & 87.1
& \cellfirst{23.6} & \cellfirst{31.0} & \cellfirst{80.5}
& \cellfirst{31.8} & \cellfirst{43.3} & \cellfirst{68.9} \\

\bottomrule
\end{tabularx}
\endgroup

        \vspace{-2pt}
        \caption{\textbf{Semantic Segmentation Probing \& Finetuning Experiments.} Chorus point-cloud variant \pts is used for ScanNet, ScanNet++, and Matterport3D, whereas the 3DGS-input model \gs is used for InteriorGS, whose source data are of 3DGS.}\label{tab:semseg_probing}
        \vspace{1mm}
    \end{minipage} \\
    \begin{minipage}[t]{0.52\textwidth}
    \centering
        \setlength{\tabcolsep}{2pt}
\renewcommand{\arraystretch}{1.05}

\begin{tabularx}{\columnwidth}{l *{8}{C}}
\toprule
Data Efficiency
& \multicolumn{4}{c}{Limited Training Scenes}
& \multicolumn{4}{c}{Limited Annotations} \\
\cmidrule(lr){1-1} \cmidrule(lr){2-5} \cmidrule(lr){6-9}
Methods
& {1\%} & {5\%} & {10\%} & {20\%}
& {20} & {50} & {100} & {200} \\
\midrule

Sonata~\cite{wu2025sonata} (lin.)
&  \cellfirst{45.3} &  \cellfirst{62.3} &  \cellsecond{68.7} & \cellsecond{69.8}
&  \cellsecond{68.8} &  \cellsecond{70.6} &  \cellsecond{71.2} &  \cellsecond{71.5} \\

{\cellcolor{ours}}\pts \ours (lin.)
& \cellsecond{42.0} & \cellsecond{60.3}
 & \cellfirst{69.6}
 & \cellfirst{71.3}
& \cellfirst{70.1} & \cellfirst{72.3}
 & \cellfirst{73.3} & \cellfirst{73.7}
 \\

\cmidrule(lr){1-9}

Sonata~\cite{wu2025sonata} (dec)
&  \cellfirst{43.8} &  \cellfirst{63.5} &  \cellsecond{69.5} &  \cellfirst{72.7}
&  \cellsecond{69.4} &  \cellfirst{72.9} &  \cellfirst{74.9} &  \cellfirst{76.3} \\

{\cellcolor{ours}}\pts \ours (dec.)
& \cellsecond{43.1} & \cellsecond{61.4} & \cellfirst{69.7} & \cellsecond{72.1}
& \cellfirst{70.8} & \cellsecond{72.4} & \cellsecond{74.1} & \cellsecond{75.3} \\

\cmidrule(lr){1-9}

PTv3~\cite{wu2024ptv3}~(sup.)
&  {25.8} &  {48.9} &  {61.0} &  {67.0}
&  {60.1} &  {67.9} &  {71.4} &  {72.7} \\

PPT~\cite{wu2024ppt}~(sup.)
&  {31.1} &  {52.6} &  {63.3} &  {68.2}
&  {62.4} &  {69.1}
&  {74.3} &  {75.5} \\

Sonata~\cite{wu2025sonata} (f.t.)
&  \cellsecond{43.5} &  \cellsecond{63.3} &  \cellsecond{71.6} &  \cellsecond{71.5}
&  \cellsecond{68.6} &  \cellsecond{72.4} &  \cellsecond{74.9} &  \cellsecond{75.9} \\

{\cellcolor{ours}}\pts \ours (f.t.)
&   \cellfirst{43.9} &   \cellfirst{64.0} &   \cellfirst{73.9} &   \cellfirst{75.0}
&   \cellfirst{73.1} &   \cellfirst{76.1} &   \cellfirst{77.2} &   \cellfirst{77.4} \\

\bottomrule
\end{tabularx}

        \vspace{-2pt}
        \caption{\textbf{\scan Data-Efficient Benchmark.}}\label{tab:semseg_data_efficiency}
        \vspace{1mm}
    \end{minipage}
    \hspace{1mm}
    \begin{minipage}[t]{0.46\textwidth}
    \centering
        \setlength{\tabcolsep}{2.0pt}
\renewcommand{\arraystretch}{1.05}

\begin{tabularx}{\columnwidth}{l *{6}{C}}
\toprule
\multirow{2}{*}[-0.7ex]{Methods}
& \multicolumn{2}{c}{ScanNet Val}
& \multicolumn{2}{c}{ScanNet200 Val}
& \multicolumn{2}{c}{ScanNet++ Val} \\
\cmidrule(lr){2-3} \cmidrule(lr){4-5} \cmidrule(lr){6-7}
& $\text{mAP}_{25}$ & $\text{mAP}_{50}$
& $\text{mAP}_{25}$ & $\text{mAP}_{50}$
& $\text{mAP}_{25}$ & $\text{mAP}_{50}$ \\

\midrule

MSC~\cite{wu2023msc} (lin.)
& 13.3 & 5.3
& 2.3 & 1.0
& 4.8 & 2.6 \\
Sonata~\cite{wu2025sonata} (lin.)
& \cellfirst{72.6} & \cellfirst{53.9}
& \cellsecond{30.0} & \cellsecond{20.9}
& \cellsecond{33.5} & \cellsecond{24.5} \\
\cellcolor{ours}\pts \ours (lin.)
& \cellsecond{66.6} & \cellsecond{46.9}
& \cellfirst{31.6} & \cellfirst{21.9}
& \cellfirst{37.0} & \cellfirst{27.9} \\

\cmidrule(lr){1-7}

Sonata~\cite{wu2025sonata} (dec.)
& \cellfirst{77.3} & \cellfirst{62.1}
& \cellsecond{36.2} & \cellsecond{29.3}
& \cellsecond{39.4} & \cellsecond{33.5} \\
\cellcolor{ours}\pts \ours (dec.)
& \cellsecond{76.9} & \cellsecond{60.5}
& \cellfirst{38.8} & \cellfirst{31.8}
& \cellfirst{41.9} & \cellfirst{33.8} \\

\cmidrule(lr){1-7}

PTv3~\cite{wu2024ptv3} (sup.)
& \textcolor{darkgray}{74.6} & \textcolor{darkgray}{57.9}
& \textcolor{darkgray}{40.1} & \textcolor{darkgray}{32.3}
& \textcolor{darkgray}{41.4} & \textcolor{darkgray}{32.5} \\
Sonata~\cite{wu2025sonata} (f.t.)
& \cellsecond{77.6} & \cellsecond{63.1}
& \cellsecond{38.3} & \cellsecond{31.5}
& \cellsecond{41.0} & \cellsecond{35.3} \\
\cellcolor{ours}\pts \ours (f.t.)
& \cellfirst{78.4} & \cellfirst{63.4}
& \cellfirst{39.3} & \cellfirst{33.7}
& \cellfirst{42.9} & \cellfirst{37.2} \\ 
\bottomrule

\end{tabularx}

        \vspace{-2pt}
        \caption{\textbf{Instance Segmentation Probing and Finetuning.}\label{tab:insseg_probing}}
        \vspace{1mm}
    \end{minipage}

\vspace{-10pt}
\end{table*}

\begin{figure}
    \centering
    \includegraphics[width=0.9\linewidth]{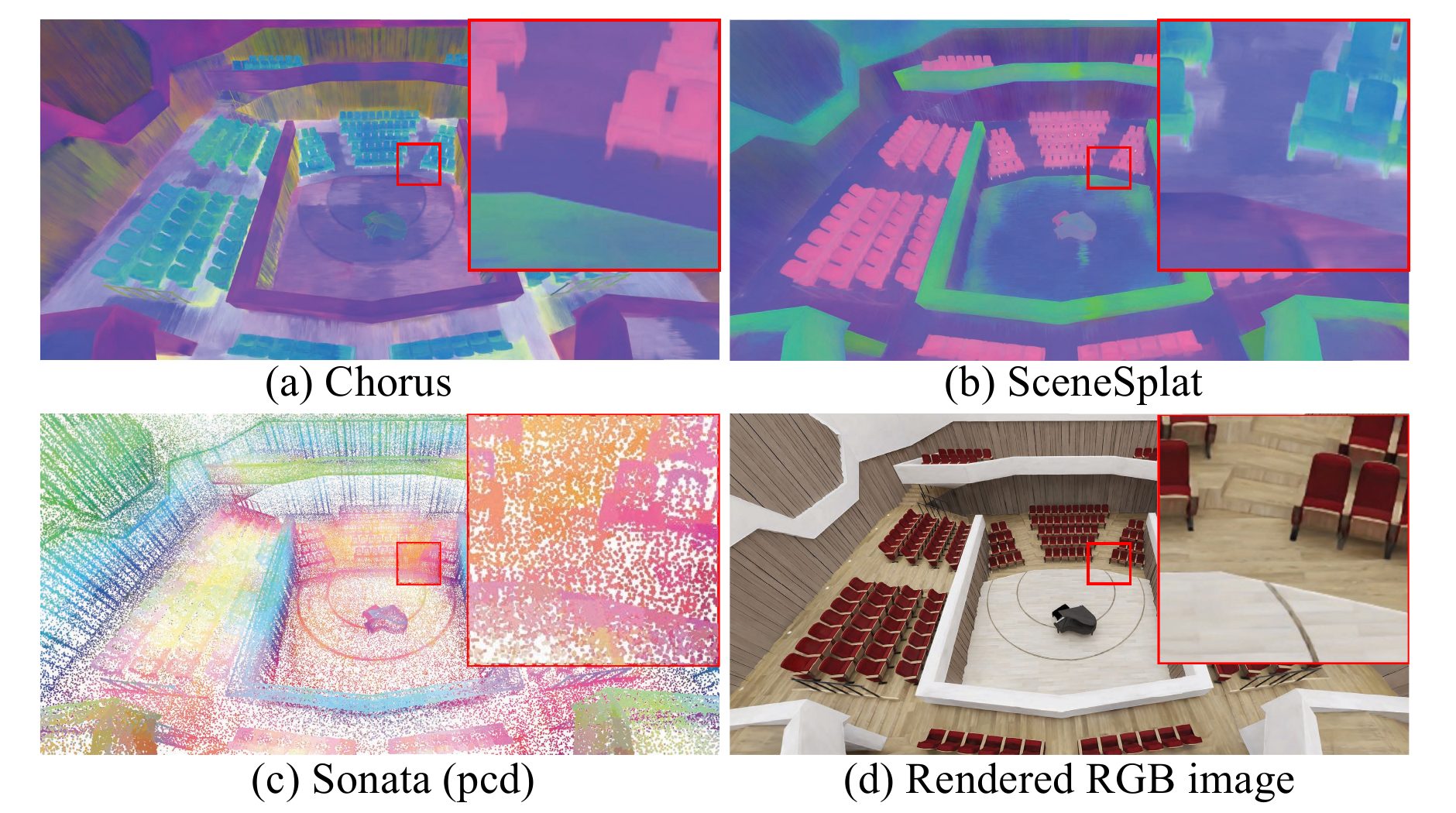}
    \caption{\textbf{Inference Feature PCA Visualization.} Features from different encoders on a concert hall. \ours shows the best semantic consistency (see zoomed-in chairs and stairs in the back).
    }
    \label{fig:feat_pca}
\end{figure}

\boldparagraph{Open-vocabulary semantic segmentation.} 
\cref{tab:semseg_zeroshot} reports zero-shot semantic segmentation on fine-grained ScanNet200, Matterport3D, ScanNet++, and \interior. \ours achieves the best zero-shot performance; compared to the previous SoTA SceneSplat, it improves by $2.1\%$ f-mIoU and $6.0\%$ f-mAcc on ScanNet200 after joint training, and by $5.7\%$ f-mIoU and $5.8\%$ f-mAcc on \interior for new-data generalization, while using $8.32\times$ less training data than the point-cloud-based pretraining method~\cite{lee2025mosaic3d}.

\boldparagraph{Open-vocabulary instance segmentation.}
We report open-vocabulary 3D instance segmentation on ScanNet200 in \cref{tab:insseg_open_vocab}. The results confirm that our encoder's strong open-vocabulary semantic understanding translates to the instance level. Following the protocol from Mosaic3D~\cite{lee2025mosaic3d}, we adopt the instance proposals from Mask3D~\cite{schult2022mask3d} for all baselines. \ours~achieves state-of-the-art performance among the methods that use 3D inputs only, outperforming prior point-cloud-based open-vocabulary SoTA~\cite{lee2025mosaic3d}. Notably, \ours~reaches a +7.6 $\text{mAP}$ gain in recognizing the 66 tail classes, showing its ability to recognize rare instances.

\begin{figure}[t]
    \centering
    \begin{subfigure}[b]{0.49\linewidth}
        \centering
        \includegraphics[width=0.9\linewidth]{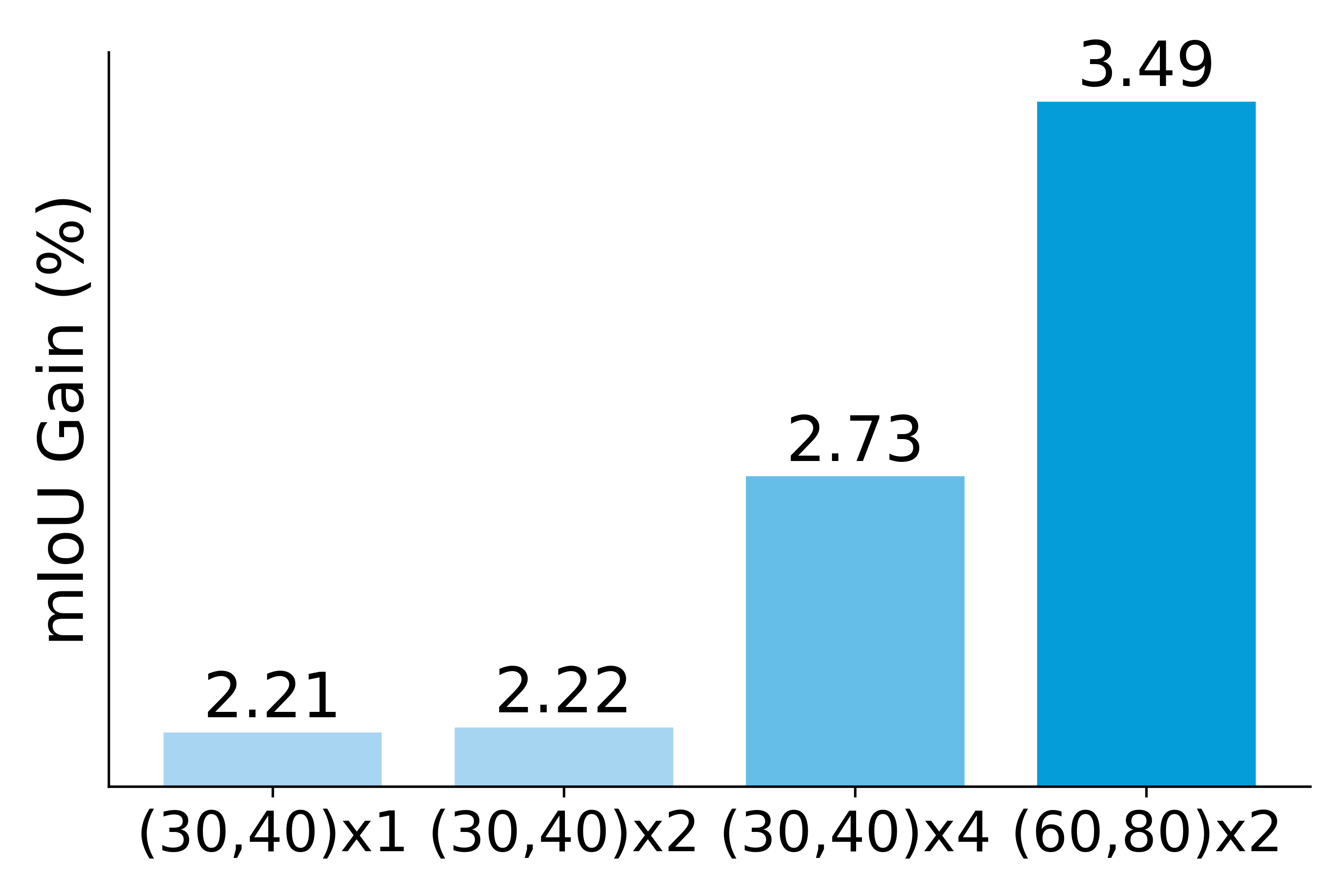}
    \end{subfigure}
    \hfill
    \begin{subfigure}[b]{0.49\linewidth}
        \centering
        \includegraphics[width=0.9\linewidth]{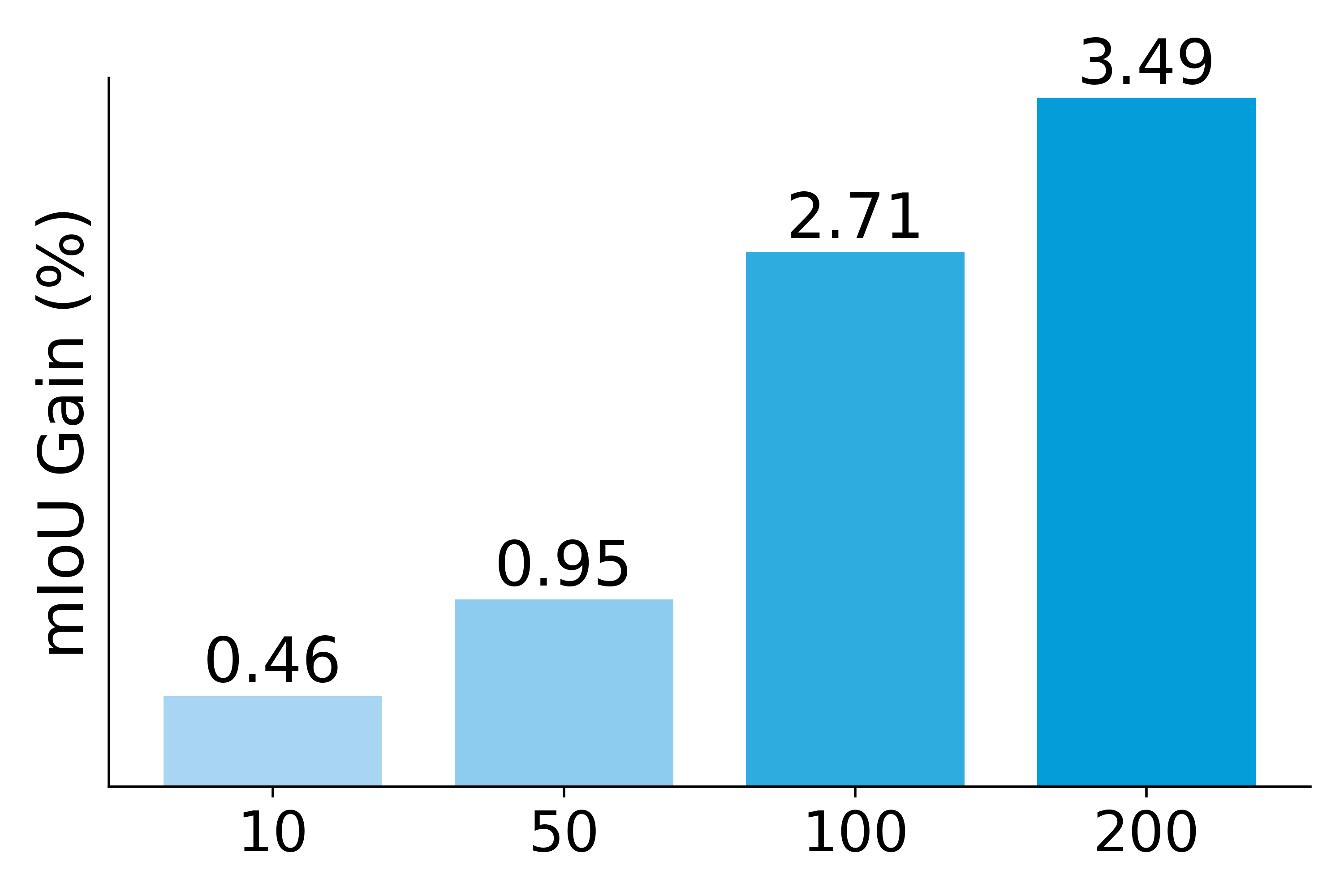}
    \end{subfigure}
    \caption{\textbf{2D Adaptation Ablation.} Performance improves with higher teacher render resolution (left) and more adaptation scenes (right). The left x-axis denotes the 2D teacher's feature resolution, formatted as (feature size) $\times$ bilinear upsample factor.}
    
    \label{fig:2D_adapt_ablation}
\vspace{-3pt}
\end{figure}

\begin{figure}
    \centering
    \includegraphics[width=0.8\linewidth]{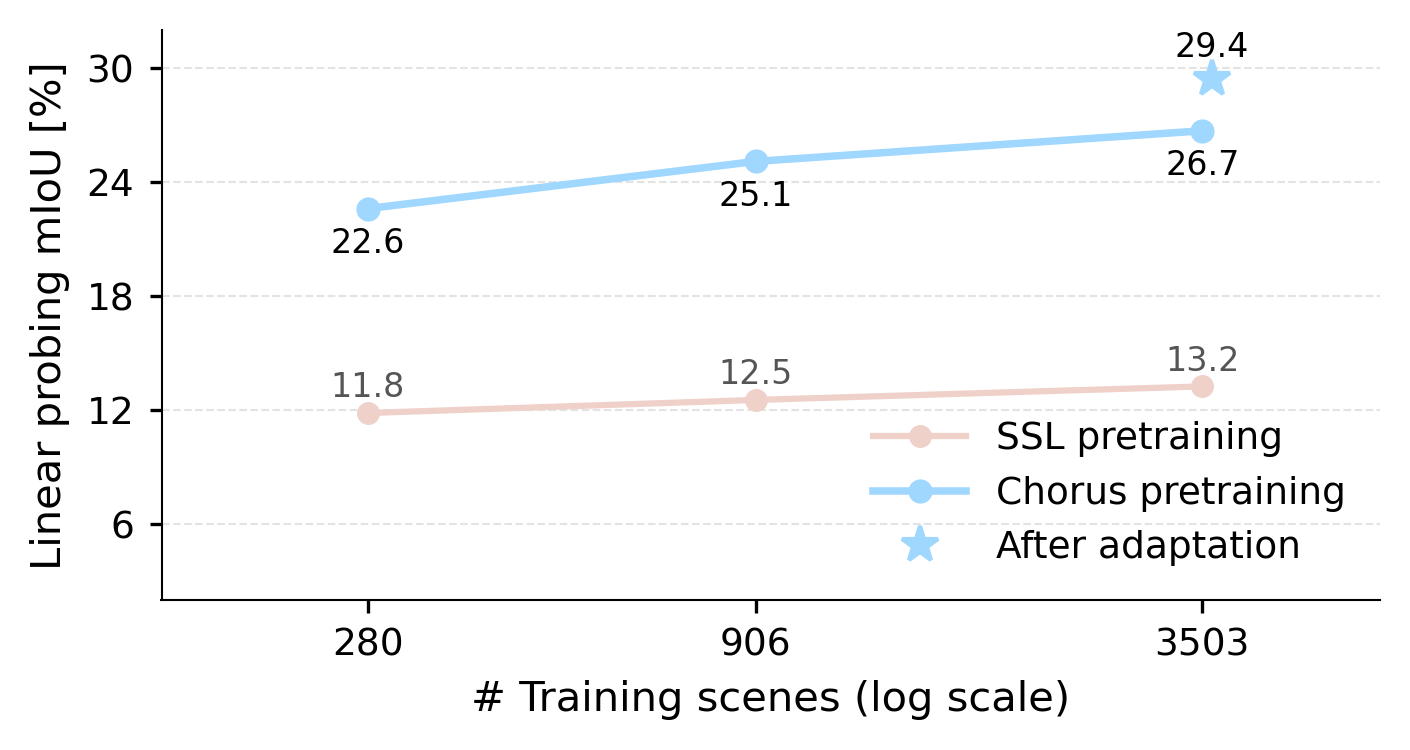}
    \caption{\textbf{Scaling Trend Together With Rendering-Based Adaptation.} Linear probing performance on \interior \vs number of pretraining scenes. We compare our multi-teacher pretraining with the self-supervised pretraining~\cite{wu2025sonata} on 3DGS; \ours scales faster and to higher accuracy. Our adaptation recipe yields a +2.7\% mIoU gain on this new dataset using only 100 scenes.}
    \label{fig:scaling_trend}
\end{figure}

\boldparagraph{Rendering-based adaptation.}
As shown in \cref{tab:2D_3D_resource}, our lightweight adaptation recipe avoids heavy feature I/O during training and adds at most $0.1$\,s per view for on-the-fly feature rasterization—eliminating the $\sim$1\,TB storage required to precompute teacher features for 800 training scenes. Its effectiveness is evident in \cref{fig:scaling_trend}: training on an additional $100$ scenes from the \interior dataset yields a $+2.7\%$ mIoU gain under linear probing over our standard pretraining, indicating better domain adaptation. We also ablate teacher feature resolution during adaptation (\cref{fig:2D_adapt_ablation}), where even low-resolution $30{\times}40$ DINOv3 features produce a clear improvement, with further gains at higher resolutions and more adaptation scenes.

\begin{table}[t]
    \centering
    \scriptsize
    \setlength{\tabcolsep}{1.8pt}
    
    \begin{tabular}{lcccc}
    \toprule
    Supervise method  & Preprocess (h) & Uplift (h) & Storage & Training Overhead \\
    \midrule
    Uplifting  & 3.4 & 2.8 & 1080 GB & - \\
    Rendering  & 0.2  & 0 & 8 GB   & Rasterization (\textless 0.1s/view) \\
    \bottomrule
    \end{tabular}
    
    \caption{\textbf{Resource and Time Comparison of Uplifting and Rendering-Based Adaptation.} Trade-off on \interior (800 scenes): preprocessing-heavy uplifting versus online-heavy rendering adaptation.}
    \label{tab:2D_3D_resource}
    
\end{table}

\begin{figure}
    \centering
    \includegraphics[width=\linewidth]{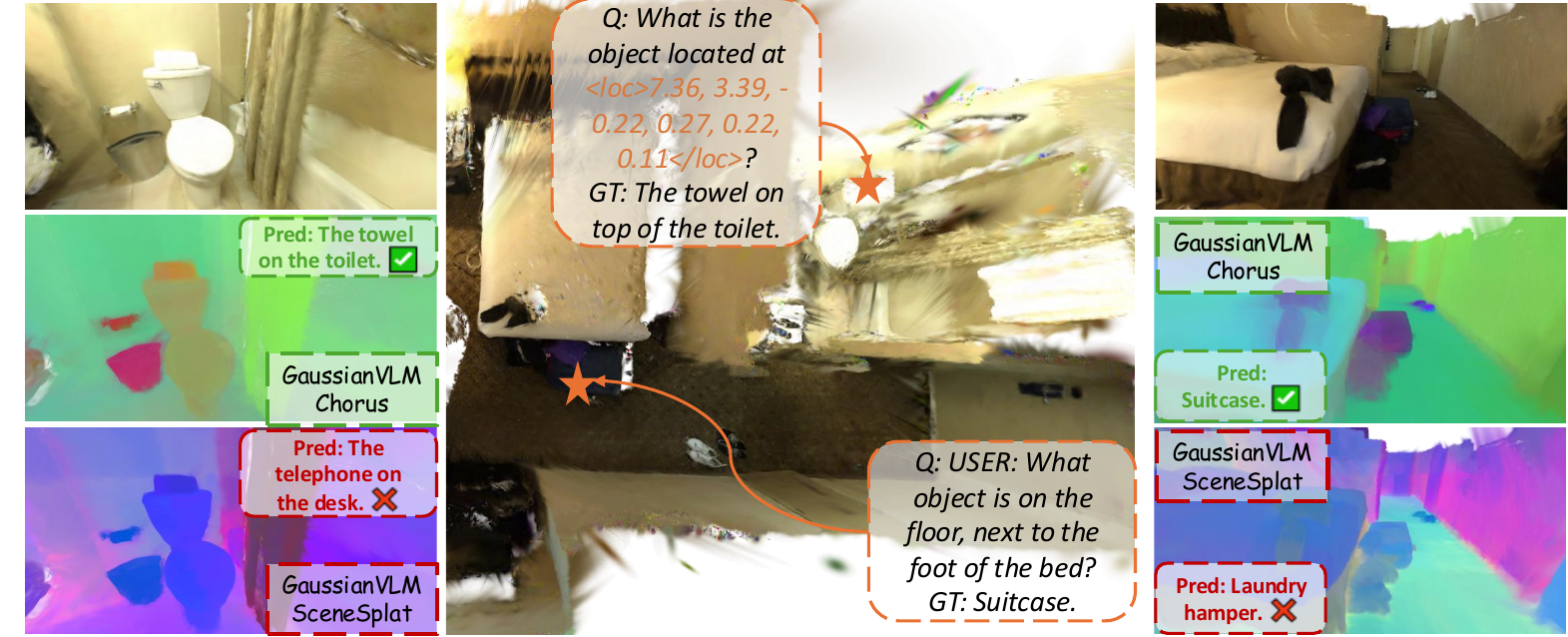}

    \caption{\textbf{VLM Qualitative Results.} We visualize a ScanNet scene with object grounding (left) and QA results (right). }
    \label{fig:scanet_vlm}
\end{figure}

\boldparagraph{Language model-based question answering.} We evaluate Chorus as the 3D encoder within an LLM-based pipeline for visual question answering and grounding (see \cref{tab:scan_all}), where swapping in \ours yields consistent improvements on both benchmarks.  Concretely, we follow GaussianVLM~\cite{halacheva2025gaussianvlm} and simply replace its 3D backbone: instead of using multi-level features from~\cite{li2025scenesplat}, we feed only the final \ours encoder stage into the VLM, keeping all other components and training settings unchanged. We train and evaluate both the original GaussianVLM and Chorus-augmented variant on ScanQA~\cite{azuma2022scanqa} (3D-VQA) and Nr3D~\cite{achlioptas2020referit3d}, using EM1 (Top-1 Exact Match), M (METEOR), R (ROUGE), and Sim (Sentence Similarity). \cref{fig:scanet_vlm} provides qualitative VQA examples.  As an additional benefit, using only the last Chorus encoder stage is lighter and faster, achieving about $0.68\times$ the training time of GaussianVLM with SceneSplat.  

\subsection{\ours on Point-Cloud Tasks}
\label{subsec:pts_results}

\boldparagraph{Probing \& finetuning of semantic segmentation.}
We evaluate the feature quality of our pretrained encoder via linear/decoder probing and full finetuning on five benchmarks, reported in \cref{tab:semseg_probing}.
With only a linear layer, \ours outperforms the strong Sonata baseline across five benchmarks, \eg, achieving mIoU gains on ScanNet200 (36.0 \vs 28.8) and ScanNet++ (48.8 \vs 40.7). When fully finetuned, \ours sets a new state-of-the-art on 4 out of 5 benchmarks, including ScanNet (79.4 mIoU) and ScanNet++ (50.2 mIoU). The advantage is particularly noticeable on ScanNet200, where \ours~achieves 40.9 mIoU, with a gain of +6.5 mIoU. Furthermore, \ours consistently achieves relatively smaller gaps between linear probing and full finetuning. These results validate that our pretraining produces separable and semantic-aware features.

\boldparagraph{Probing \& finetuning of instance segmentation.}
We extend analysis to instance segmentation in \cref{tab:insseg_probing}. Linear probing again shows the strength of our features; while Sonata leads on ScanNet, \ours outperforms it on ScanNet200 (31.6 $\text{mAP}_{25}$) and ScanNet++ (37.0 $\text{mAP}_{25}$). When fully finetuned, \ours remains competitive, achieving the best results on ScanNet++ (42.9 $\text{mAP}_{25}$) and performing comparably to top supervised methods on ScanNet.

\boldparagraph{Data efficiency experiments.}
We validate the benefit of our pretraining under data-scarce conditions on ScanNet in \cref{tab:semseg_data_efficiency}. The results show our encoder's pretrained features provide advantages over the Sonata baseline. When fully finetuned, \ours~consistently outperforms Sonata across all limited-scene (1\%-20\%) and limited-annotation (20-200 points/scene) settings. This demonstrates that our pretraining particularly helps in the low-data regime (\eg, +4.5 mAP with 20 labels). 

\subsection{Ablation and Analysis}
\label{subsec:ablation}

\boldparagraph{Why does \ours work well on point clouds?}
We examine the Chorus variant that uses only Gaussian centers, colors, and normals as input while keeping the multi-teacher objectives unchanged. Despite the distribution gap between point-cloud observations and optimized 3DGS parameters, we test two hypotheses: \textit{(i)} 3DGS pretraining acts as a strong point-cloud augmentation, inducing noise-robust features; \textit{(ii)} multi-teacher pretraining is more data-efficient than self-supervised pretraining, yielding better scaling.

For \textit{(i)}, we retrieve instance-level features from original point clouds (PC) to perturbed PC (centers with Gaussian noise). We report $\mathrm{R@1}$—the fraction of top-1 matches from the same instance—and \emph{Same-class@Incorrect top-1}, which measures how often an incorrect match is at least the correct semantic class. Evaluated using 684 instances from 10 \ppv2 Val scenes, the Chorus variant performs better on both (Tab.~\ref{tab:ablation_retrieval}), indicating robustness to input noise. For \textit{(ii)}, InteriorGS linear probing shows that \ours scales faster than the self-supervised Sonata scheme as pretraining scenes increase (\cref{fig:scaling_trend}). Together, these results suggest that 3DGS pretraining induces noise-robust embeddings, while multi-teacher supervision supplies scaling signals that explain the variant’s strong PC performance. See \cref{fig:feat_pca} and the supplement for PCA visualizations.

\begin{figure}
    \centering
    \includegraphics[width=\linewidth]{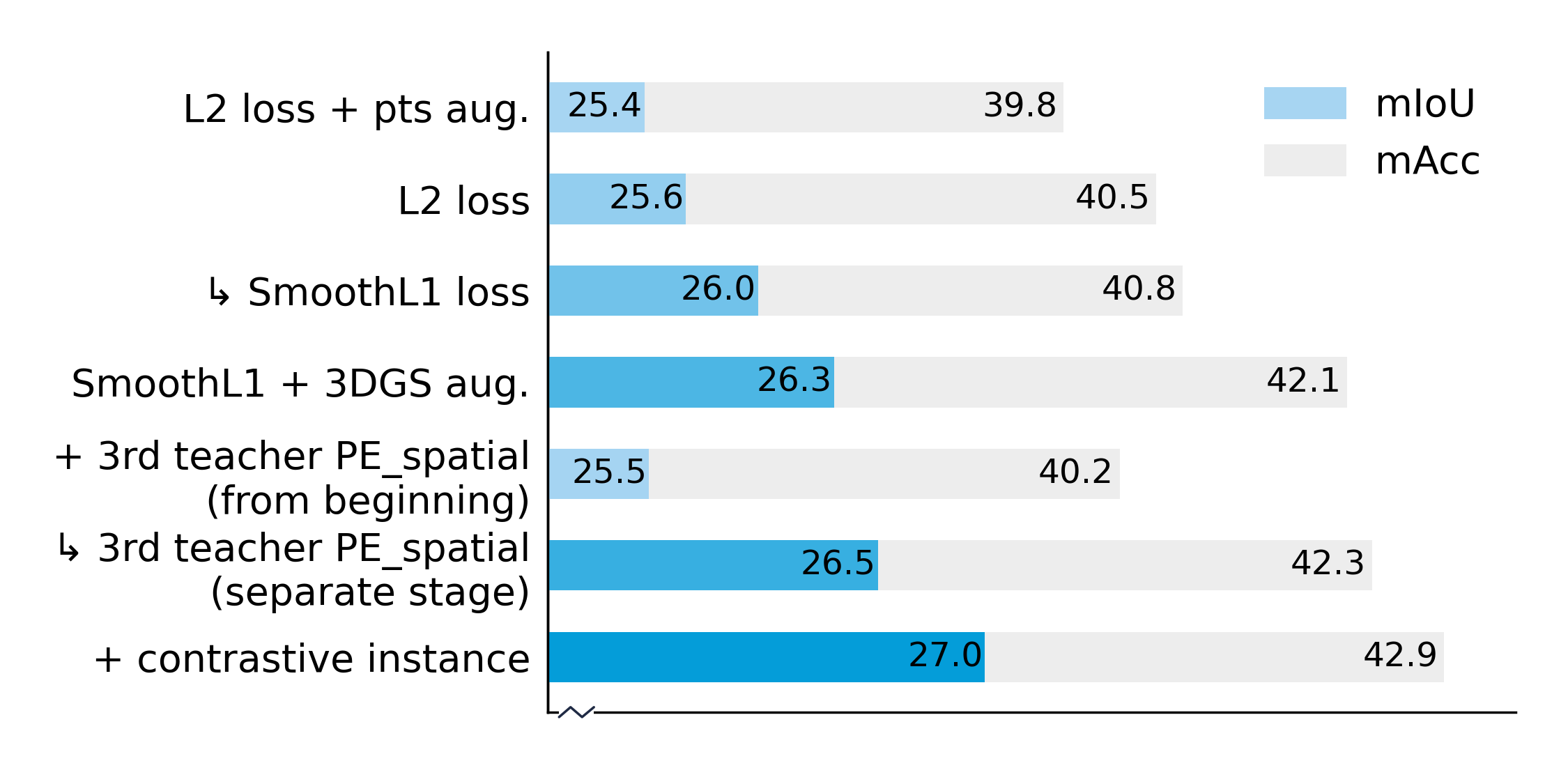}
    \vspace{-6mm}
    \caption{\textbf{Design Choice Ablation.} We validate the choices by evaluating zero-shot segmentation on \ppv2 Val using a subset of training scenes. SmoothL1 loss, 3DGS-aware augmentations, introducing PE-Spatial in a separate stage, and an instance-level contrastive term each provide incremental gains.}
    \label{fig:ablation_design_choice}
\end{figure}

\begin{table}[t]
\centering
\scriptsize
\setlength{\tabcolsep}{4pt}
\renewcommand{\arraystretch}{1.0}

\begin{tabular}{lcc}
\toprule
Method & $\mathrm{R@1}$ (PC$\!\to$noisy PC)$\uparrow$ & Same-class@Incorrect top-1$\uparrow$ \\
\midrule
Sonata & 79.8\% & 75.0\% \\
{\cellcolor{ours}}\ours variant & \cellfirst{85.4\%} & \cellfirst{78.0\%} \\
\bottomrule
\end{tabular}
\caption{\textbf{Instance Retrieval From PC to Perturbed PC.} Averaged over 684 instances from 10 ScanNet++ Val scenes.}

\label{tab:ablation_retrieval}
\end{table}

\begin{table}[ht]
\centering
\footnotesize
\setlength{\tabcolsep}{2.2pt}
\renewcommand{\arraystretch}{1.05}

\begin{tabularx}{\linewidth}{@{}L{20mm} ccc *{4}{C}}
\toprule
\multirow{2}{20mm}[-0.7ex]{\raggedright Training\\source} & \multicolumn{3}{c}{Teachers} & \multicolumn{2}{c}{Val Split} & \multicolumn{2}{c}{InteriorGS} \\
\cmidrule(lr){2-4} \cmidrule(lr){5-6} \cmidrule(lr){7-8}
& Lang & DINO & PE & mIoU\(_\text{fg}\) & mAcc\(_\text{fg}\) & mIoU\(_\text{fg}\) & mAcc\(_\text{fg}\) \\
\midrule

\multirow{2}{*}{ScanNet} & \checkmark & – & – & 21.2 & 42.0 & 7.3 & 8.8 \\
                & \checkmark & \checkmark & – & \cellfirst{22.4} & \cellfirst{45.8} & \cellfirst{9.0} & \cellfirst{14.6} \\

\cmidrule{1-8}

\multirow{3}{*}{ScanNet++ v2}  & \checkmark & – & – & 27.1 & 45.3 & 8.0 & 12.8 \\

& \checkmark & \checkmark & – & 29.4 & 55.8 & 9.3 & 16.0 \\

& \checkmark & \checkmark & \checkmark & \cellfirst{29.6} & \cellfirst{56.4} & \cellfirst{11.4} & \cellfirst{17.1} \\

\bottomrule
\end{tabularx}

\caption{\textbf{Teacher Ablation with Zero-Shot Semantic Segmentation.} The “Teachers” columns mark included components (\checkmark/–). We report foreground metrics for all settings.}
\label{tab:ablation_teachers}
\end{table}

\boldparagraph{Analysis on multi-teacher objective.}
Although \ours has multiple supervision, recent evidence suggests pretrained vision models with diverse training objectives still share non-trivial alignable subspaces 
(representational convergence~\cite{platonic2024}; shared units across supervision~\cite{dravid2023rosetta}).
Concretely, our teachers are heterogeneous but not arbitrary: SigLIP2 and PE both target contrastive language-alignment, while DINOv3 relies on student-teacher distillation. All are trained on billion-scale web images, leading to potential overlap in mid-level structure (objectness/parts/correspondence) despite different emphases. We exploit this by learning a single shared 3DGS backbone for the common factors, while using lightweight per-teacher projectors to absorb teacher-specific embedding idiosyncrasies, so objectives need not compete in one output space. Lastly, the PHI-S balancing we adopted further reduce dominant interference from teacher features. 

\boldparagraph{Teacher ablation.}
We ablate the three teachers—SigLIP, DINO, and PE‐Spatial—in two complementary views. In~\cref{tab:ablation_teachers}, adding DINO and then PE to the language teacher consistently improves zero‐shot semantic segmentation on both the training dataset and InteriorGS, indicating complementary semantics (Lang) and general, object‐aware structure (DINO/PE). We further ablate the SigLIP teacher in the supplement. Together, the results show non-redundant gains from each teacher.

\boldparagraph{Robustness to 3DGS variants.} 
In~\cref{tab:rebuttal_robustness}, we evaluate \ours (\gs) on scenes re-optimized with standard 3DGS (without MCMC~\cite{kheradmand20243d}) and on scenes re-optimized using MCMC with fewer Gaussians. Zero-shot segmentation performance is largely preserved, suggesting that the learned encoder is not sensitive to the optimization algorithm or reduced Gaussian density.

\begin{figure}[h]
    \centering
    \begin{minipage}[t]{0.55\linewidth}
        \vspace{0pt}
        \centering
        \scriptsize
        \setlength{\tabcolsep}{1.0pt}
        \renewcommand{\arraystretch}{1.0}
        \begin{tabularx}{\linewidth}{l|CC}
            \toprule
            f-mIoU/f-mAcc(\%) & ScanNet200 & \ppv2 \\
            \midrule
            MCMC (ours) & 24.6\,/\,47.7 & 29.6\,/\,53.5 \\
            Standard    & 22.8\,/\,42.6 & 29.0\,/\,48.6 \\
            \bottomrule
        \end{tabularx}
    \end{minipage}
    \hfill
    \begin{minipage}[t]{0.40\linewidth}
        \vspace{0pt}
        \centering
        \scriptsize
        \setlength{\tabcolsep}{1.5pt}
        \renewcommand{\arraystretch}{1.0}
        \begin{tabularx}{\linewidth}{l| CCC}
            \toprule
            Total GS & 10\% & 50\% & 100\% \\
            \midrule
            f-mIoU & 27.0 & 29.5 & 29.6 \\ 
            f-mAcc & 46.6 & 53.1 & 53.5 \\ 
            \bottomrule
        \end{tabularx}
    \end{minipage}
    \captionof{table}{\textbf{Robustness to 3DGS Optimization and Density.}}
    \label{tab:rebuttal_robustness}
\end{figure}

\boldparagraph{Robustness to iPhone RGB-D input.} 
We probe \ours variant (\pts) on \ppv2 iPhone-captured point clouds (\vs from laser scans), and observe consistent performance in~\cref{tab:rebuttal_probing_lidar}, suggesting robustness in point clouds encoding.

\begin{table}[ht]
    \centering
    \scriptsize
    \setlength{\tabcolsep}{2.5pt}
    \renewcommand{\arraystretch}{1.0}
        \begin{tabularx}{\columnwidth}{l *{3}{M{18mm}}}
        \toprule
         SN++ Linear Probing & Laser Scan & iPhone & iPhone \\
        \midrule
        mIoU/mAcc (\%)  & \ours 48.8\,/\,63.2 & \ours 45.7\,/\,57.6 &  Sonata 39.5\,/\,52.0 \\
        \bottomrule
        \end{tabularx}
    \captionof{table}{\textbf{Robustness to iPhone RGB-D Captures.}}
    \label{tab:rebuttal_probing_lidar}
\end{table}

\boldparagraph{Design choice ablation.}
~\cref{fig:ablation_design_choice} evaluates training choices on \ppv2 Val using a subset. SmoothL1 loss, 3DGS‐aware augmentations, and an instance‐level contrastive term each yield incremental improvements. Staging PE-Spatial only in the second half of training works best: early PE may over-anchor to local features, whereas late application refines spatial awareness after a stable backbone has formed.

\begin{table}[ht!]
    \centering
    \scriptsize
    \begin{tabular}{lcccc} 
    
    \toprule
    Method & \gs SceneSplat  & {\cellcolor{ours}}\gs \ours & Sonata & Mosaic3D \\
    \midrule
    
    \#Model Params.        & 91.7M   & 131.3M & 108.5M  & 39.1M  \\
    Inference Time/Scene  & 0.65s   &  0.70s  & 0.49s  & 0.25s  \\
    
    \bottomrule
    \end{tabular}
    
    \caption{\textbf{Model Size \& Runtime.} Averaged on 100 scenes.}
    
    \label{tab:ablation_runtime}
\end{table}

\boldparagraph{Runtime.} \cref{tab:ablation_runtime} compares model size and average inference time on 100 \interior test scenes (965K Gaussians on average). \ours is the slowest but remains practical at 0.7s per scene.

\section{Conclusion}
\label{sec:conclusion}
We introduced \ours, a multi-teacher pretraining framework that learns general-purpose 3D scene representations directly from 3D Gaussian splats. By aligning a native 3DGS encoder with complementary 2D foundation models, \ours distills language-aligned, generalist, and spatially local cues into a unified 3D embedding that transfers well across scene understanding tasks. Extensive experiments on 3DGS-native and point-cloud benchmarks show state-of-the-art performance and efficient render-and-distill adaptation to new domains. A remaining limitation is the offline cost of precomputing teacher pseudo-labels, and an interesting direction is to move toward a unified point-cloud–3DGS encoder built on our findings.

\section{Acknowledgement} 
Yue Li is financially supported by TomTom, the University of Amsterdam, and the allowance from the Top consortia for Knowledge and Innovation (TKIs) from the Netherlands Ministry of Economic Affairs and Climate Policy. This work used the Dutch national e-infrastructure with the support of the SURF Cooperative under grant no. NWO-2024.035. We thank Caspar van Leeuwen and the SURF team for their invaluable support with the compute cluster. This work was partially supported by INSAIT, Sofia University “St. Kliment Ohridski”, the MUR PNRR project FAIR (PE00000013), and the EU Horizon projects ELIAS (No. 101120237) and ELLIOT (No. 10121439).

{
    \small
    \bibliographystyle{ieeenat_fullname}
    \bibliography{main}
}

\clearpage
\setcounter{page}{1}
\maketitlesupplementary

\setcounter{section}{0}
\setcounter{figure}{0}    
\setcounter{table}{0}    
\renewcommand{\thetable}{\Alph{table}}
\renewcommand{\thefigure}{\Alph{figure}}
\renewcommand{\thesection}{\Alph{section}}

\renewcommand{\theHsection}{supp.\Alph{section}}
\renewcommand{\theHsubsection}{supp.\Alph{section}.\arabic{subsection}}
\renewcommand{\theHsubsubsection}{supp.\Alph{section}.\arabic{subsection}.\arabic{subsection}}
\renewcommand{\theHfigure}{supp.\Alph{figure}}
\renewcommand{\theHtable}{supp.\Alph{table}}
\makeatletter
\@ifundefined{theHalgorithm}{}{\renewcommand{\theHalgorithm}{supp.\arabic{algorithm}}}
\makeatother

\addtocontents{toc}{\protect\setcounter{tocdepth}{2}}

\definecolor{customblue}{HTML}{215FDB} %
{
\setcounter{tocdepth}{2}   %
\hypersetup{linkcolor=customblue}
\tableofcontents
}


\section{Method Details}
\label{supp:method}

\subsection{Teacher-Specific Contrastive Loss}
\label{supp:contrastive}

When the source dataset for a teacher $t$ provides semantic or instance labels, we optionally add a compact contrastive regularizer $\mathcal{L}_{\text{con}}^{(t)}$ on top of the per-Gaussian matching losses. This follows the aggregated InfoNCE formulation of SceneSplat~\cite{li2025scenesplat}, but is applied independently per teacher and per supervision type (semantic \& instance).

Let $\hat{f}^{(t)} \in \mathbb{R}^{N \times d_t}$ denote the predicted per-Gaussian features for teacher $t$, where $N$ is the number of Gaussians in the scene. We assume that the Gaussians are partitioned into a collection of labeled groups $\{\mathcal{G}_g\}_{g \in \mathcal{C}^{(t)}}$, where $g$ indexes either a semantic class (for SigLIP2) or an instance (for PE-Spatial), and $\mathcal{C}^{(t)}$ is the set of valid groups for teacher~$t$.

For each group $g \in \mathcal{C}^{(t)}$ with sufficiently many Gaussians, we randomly split its elements into two disjoint subsets
\(
\mathcal{G}_g^A, \mathcal{G}_g^B \subset \mathcal{G}_g
\)
and compute pooled features
\begin{equation}
\bar{f}_g^A = \text{mean}\{\hat{f}^{(t)}_i : i \in \mathcal{G}_g^A\},
\;
\bar{f}_g^B = \text{mean}\{\hat{f}^{(t)}_i : i \in \mathcal{G}_g^B\}.
\end{equation}
Stacking these vectors over all groups yields matrices
\(
\bar{f}^A, \bar{f}^B \in \mathbb{R}^{|\mathcal{C}^{(t)}| \times d_t}
\),
which we $\ell_2$-normalize row-wise and use to form bidirectional InfoNCE logits:
\begin{equation}
Z^A = \bar{f}^A (\bar{f}^B)^\top / \tau^{(t)},
\;
Z^B = \bar{f}^B (\bar{f}^A)^\top / \tau^{(t)},
\end{equation}
where $\tau^{(t)}$ is a learnable temperature. Each row of $Z^A$ and $Z^B$ corresponds to a query group, and the diagonal entries are the positive pairs (same group across the $A/B$ split). The teacher-specific contrastive loss is then
\begin{align}
\mathcal{L}_{\mathrm{con}}^{(t)}
&= \frac{1}{2|\mathcal{C}^{(t)}|}
\sum_{X \in \{A,B\}}
\sum_{g \in \mathcal{C}^{(t)}}
-\log
\frac{\exp\big(Z^X_{g,g}\big)}
{\sum_{g' \in \mathcal{C}^{(t)}} \exp\big(Z^X_{g,g'}\big)}.
\label{eq:teacher_contrastive}
\end{align}

We use the same formulation in Eq.~\eqref{eq:teacher_contrastive} for different supervision signals:

\begin{itemize}
\item \textbf{SigLIP2 (semantic).}
Here, each group $g$ is a semantic class $c$, and $\mathcal{G}_c$ is the set of Gaussians assigned to it.
We thus pool class-wise features
$
\bar{f}_c^A, \bar{f}_c^B
$
from SigLIP2-predicted features $\hat{f}^{(t)}$ and apply Eq.~\eqref{eq:teacher_contrastive} over $\mathcal{C}^{(t)} = \{c\}$.

\item \textbf{PE-Spatial (instance).}
Here, each group $g$ is an instance $k$, and $\mathcal{G}_k$ collects Gaussians belonging to it.
We pool instance-wise means
$
\bar{f}_k^A, \bar{f}_k^B
$
and apply the same loss over $\mathcal{C}^{(t)} = \{k\}$, encouraging distinct instances to occupy well-separated regions in the embedding space.
\end{itemize}

In both cases, $\mathcal{L}_{\mathrm{con}}^{(t)}$ is enabled only after the warm-up epochs of the teacher projector so that the features first stabilize under the regression objectives before being refined by contrastive separation.

\subsection{Rendering-Based Adaptation}

\label{supp:2d_adapt_method_detail}
\begin{figure}[t]
    \centering
    \includegraphics[width=\linewidth]{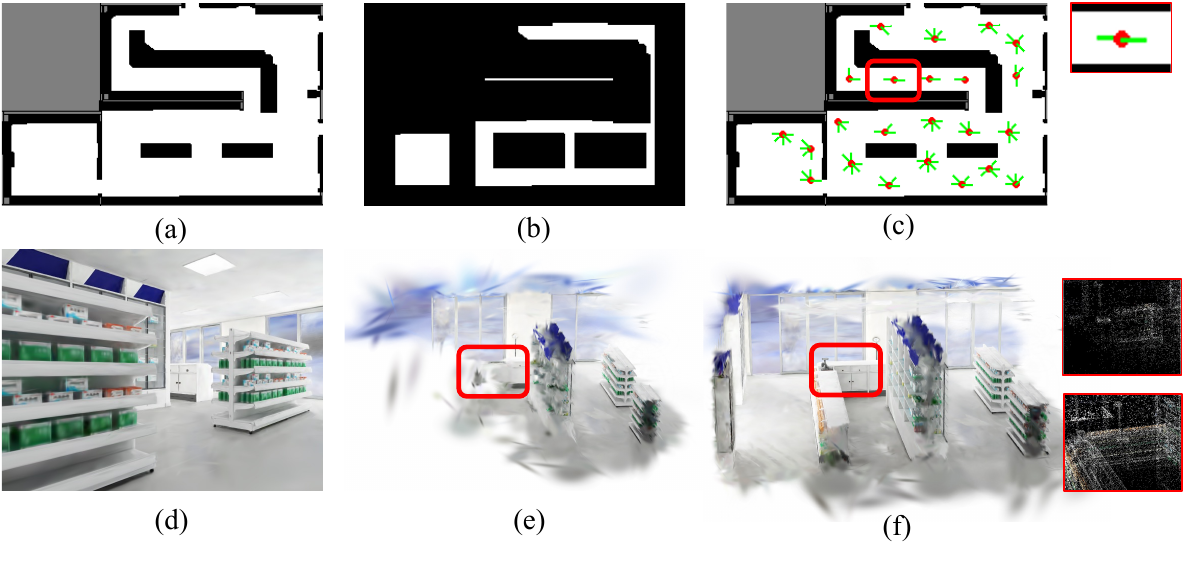}
    \caption{\textbf{Visualization of Pose Sampling and Visibility Crop.} (a) The 2D occupancy map. (b) The dilated occupancy map. (c) The sampled locations and view directions, with a zoomed-in view showing that viewpoints near obstacles are excluded. (d) The 2D rendered image for the given pose. (e) The naive visibility crop for the given pose. (f) The box-augmented visibility crop, with a zoomed-in image showing that more complete 3D structure is maintained.}
    \label{fig:2d_adapt_vis}
\end{figure}

We further explain the pose sampling and selection pipeline. The primary requirement for sampling camera poses is finding valid locations within free space. For Gaussian Splats, this can be achieved by computing the signed distance field through rendered-depth-map fusion. For convenience, we leverage the pre-computed 2D occupancy maps provided by the \interior dataset. To avoid poses that are too close to obstacles (where the occlusion map may contain errors and the resulting view is less informative), we first dilate the occlusion mask as shown in \cref{fig:2d_adapt_vis}. Next, we sample the 2D camera locations within the dilated occupancy map. To address cases where the occlusion map incorrectly marks areas outside the room as free space, we ignore all poses that fall outside the $XY$ projection boundary of the Gaussian splats. 

Determining the optimal camera height (Z-coordinate) is non-trivial, as simply using the mean or median height ignores variations like those found in multi-story environments. We propose a label-guided height decision method: for a sampled $XY$ position, we query labeled points within a 2-meter radius, prioritizing furniture labels, then floor labels, and finally ceiling labels. We use the height of these labeled points to determine the camera's Z-location (specifically, $\pm 0.5\text{m}$ relative to furniture, $+1.5\text{m}$ relative to the floor, and $-1\text{m}$ relative to the ceiling), thereby avoiding viewpoints that are either too low or too high.

We sample eight uniform view directions from the given location. For each direction, we project the center ray onto the XY plane and check if this ray intersects any obstacles within a 2-meter radius. If an intersection occurs, we disregard that direction. As illustrated in \cref{fig:2d_adapt_vis}, only view directions parallel to the narrow corridor are retained.

\begin{figure}[t]
    \centering
    \centering
    \begin{subfigure}[b]{0.24\linewidth}
        \centering
        \includegraphics[width=\linewidth]{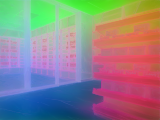}
        \caption{}
    \end{subfigure}
    \hfill
    \begin{subfigure}[b]{0.24\linewidth}
        \centering
        \includegraphics[width=\linewidth]{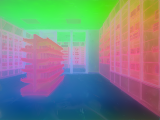}
        \caption{}
    \end{subfigure}
    \hfill
    \begin{subfigure}[b]{0.24\linewidth}
        \centering
        \includegraphics[width=\linewidth]{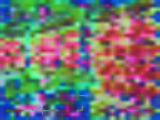}
        \caption{}
    \end{subfigure}
    \hfill
    \begin{subfigure}[b]{0.24\linewidth}
        \centering
        \includegraphics[width=\linewidth]{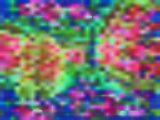}
        \caption{}
    \end{subfigure}
    \caption{\textbf{Cross-View 2D Teacher Feature Visualization.} PCA visualizations of DINOv3 features in (a) View 1 and (b) View 2, compared to SigLIP2 features in (c) View 1 and (d) View 2. DINOv3 generally demonstrates good 3D consistency across views. In contrast, SigLIP2 patch features are less smoothed and consistent across views.}
    \label{fig:frame_consistency}
\end{figure}

Next, we determine the visible Gaussians by selecting the splats that have valid accumulated transmittance during rasterization. As illustrated in \cref{fig:2d_adapt_vis}, this visibility is linked primarily to the 2D rendering rather than the underlying 3D scene structure. If we supervise the network using only these visible splats, critical geometry (such as the desk structure) can be missed. To address this, we compute the minimum 2D bounding box that encloses all visible splats and augment it with an additional $0.2\text{m}$ margin. This results in a complete and larger scene region for the subsequent Chorus encoder forward pass.

The enclosed splats within the augmented bounding box are saved. Next, we find three additional poses from the remaining pose set that exhibit the largest overlap ratio with this specific bounding box. These selected cameras are then used in the subsequent training step. This strategy allows us to perform a forward pass on a single cropped 3D scene and supervise it using up to four images from different viewpoints, effectively promoting inter-view encoding consistency, as analyzed in \cref{fig:frame_consistency}.

\begin{figure}
    \centering
    \includegraphics[width=\linewidth]{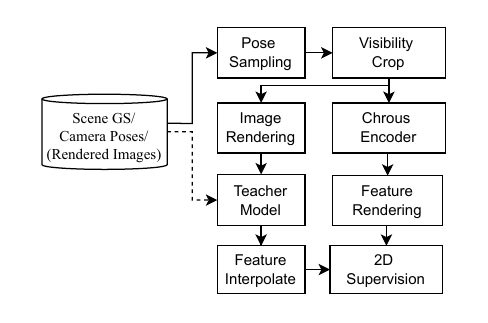}
    \caption{\textbf{Rendering-Based Adaptation Diagram.}}
    \label{fig:2d_diagram}
\end{figure}

Following the above preprocessing, we present the complete training pipeline. \cref{fig:2d_diagram} shows the high-level workflow and \cref{alg:2d_adaption_train} provides the details. For each batch, we first select multiple scenes. Within each selected scene, we sample one main camera pose and then sample $n-1$ additional camera poses for rendering supervision. We then merge the visibility masks from these poses to form a co-visibility mask. As demonstrated in \cref{tab:2d_data_summary}, our overlap ratio ranking strategy helps prevent the inclusion of excessive Gaussians during merging.

The pipeline then proceeds in two parallel branches. First, we render the co-visible Gaussians to synthesize a batch of images for each pose (or utilize pre-rendered views). These images are fed into the 2D teacher model to extract features, which are subsequently interpolated to the target resolution. Simultaneously, the co-visible Gaussians are forwarded to the Chorus encoder to obtain per-Gaussian features. These per-Gaussian features are then rendered back onto the 2D plane to obtain the predicted 2D feature maps ${\hat{F}}_j$ for the sampled poses. Finally, the loss is computed between the predicted features ${\hat{F}}_j$ and the teacher features ${\tilde{F}}_j$.

 \begin{algorithm}[t]
    \caption{Rendering-Based Adaptation}
    \begin{algorithmic}[1]
    \State \textbf{Input:} Gaussian scene $G=\{G_i\}_{i=1}^N$, training poses $\{P_j\}_{j=1}^M$ and views $\{I_j\}_{j=1}^M$ (optional) of this scene, overlap set size $n$, and pose visibility lookup table $\{V_j\}_{j=1}^M$.
    \State \textbf{Models:} 2D teacher model $t$ and Chorus encoder $g_\theta$.
    \State \textbf{Output:} Loss for Rendering-Based Adaptation.
    \State \textbf{Step 1: Pose Sampling and Scene Crop}
    \State Uniformly sample $P_s \sim \mathcal{U}(\{P_j\}_{j=1}^M)$ 
    \State Given the lookup table, calculate overlap counts between $P_s$ and all other poses $P_g$:
    $$ E(P_s, P_g) = | V_s \cap V_g | \quad \forall P_g \in \mathcal{P}, g \neq s $$
    \State Define selection probability $\mathbf{Prob}(P_g)$ proportional to overlap:
    $$ \mathbf{Prob}(P_g) = \frac{E(P_s, P_g)}{\sum_{k \neq s} E(P_s, P_k)} $$
    \State Sample $n-1$ overlap poses $\text{Pair}_s$ to form the batch $\mathcal{B}$:
    $$ \mathcal{B} \leftarrow \{P_s\} \cup \text{Sample}(\mathcal{P} \setminus \{P_s\}, n-1, \mathbf{Prob}) $$
    \State \textbf{Step 2: Visibility Crop}
    \State Create the co-visibility mask of the batch with $\mathbf{V} = \bigcup_{V_i \in \mathcal{B}} V_i$, and then crop the scene ${G'} = \text{Crop}(G,\mathbf{V}) $
    \State \textbf{Step 3: Image Rendering}
    \For{$P_j \in \mathcal{B}$}
        \State Obtain RGB image $\tilde{I}_j$ (Render $G'$ via rasterization or retrieve pre-rendered $I_j$).
    \EndFor
    \State \textbf{Step 4: Chorus Encoding}
    \State Encode the cropped scene $G'$ to get $g_\theta({G'})$.

    \State \textbf{Step 5: Teacher Supervision}
    \For{$P_j \in \mathcal{B}$}
        \State \textbf{Teacher:} Run inference and extract teacher 2D features $\tilde{F}_j^{(t)}$ from image $\tilde{I}_j$.
        \State \textbf{Chorus:} Render predicted feature map using $g_\theta({G'})$:
        $$ \hat{F}_j = \text{Render}(G', g_\theta({G'}), P_j) $$, with resolution $H_{\hat{F}}, W_{\hat{F}}$
    \EndFor

    \State \textbf{Step 6: Loss Calculation}
    \State To align the features, we interpolate the teacher feature resolution to $H_{\hat{F}}$ and calculate the loss $\mathbf{L} = \mathcal{L}(\hat{F}_{j}^{(t)}, \text{Interp}(\tilde{F}_j^{(t)}))$.

    \State \textbf{Return:} Rendering-Based Adaptation loss $\mathbf{L}$
    \end{algorithmic}
    \label{alg:2d_adaption_train}
\end{algorithm}

\subsection{3DGS-Aware Augmentations}
\label{supp:augs}

We now detail the two 3DGS-aware augmentations introduced in the main paper, which are designed to perturb splat parameters in ways that are approximately rendering-preserving and consistent with observed 3DGS optimization dynamics.

\boldparagraph{Rendering-equivalent perturbation.}
Let $R(\mathbf{q}_i)\!\in\!\mathrm{SO}(3)$ be the rotation matrix of quaternion $\mathbf{q}_i$, and the world-space covariance of the $i$-th splat be
\begin{equation}
\Sigma_i \;=\; R(\mathbf{q}_i)\,\mathrm{diag}(\mathbf{s}_i^2)\,R(\mathbf{q}_i)^\top.
\end{equation}
Motivated by \emph{exploration steps} in 3DGS-MCMC~\cite{kheradmand20243d}, where low-contribution splats are perturbed or relocated without harming the converged quality, we add small, zero-mean displacements to the ellipsoid centers and modulate their magnitude by the splat’s opacity. Concretely, for each splat we sample
\begin{equation}
\label{eq:noise}
\mathbf{x}_i' \;=\; \mathbf{x}_i \;+\; \eta \, w(\alpha_i)\, \Sigma_i \boldsymbol{\xi}_i,
\quad \boldsymbol{\xi}_i \sim \mathcal{N}(\mathbf{0}, I).
\end{equation}
where $\eta\!>\!0$ is a small scale and $w(\alpha)$ is a monotonic weight emphasizing \emph{low-opacity} (i.e., less influential or certain) splats. In practice, we use the sharp logistic function as in~\cite{kheradmand20243d}
\begin{equation}
w(\alpha)
=
\sigma\!\big(k(\tau - \alpha)\big),
\quad
k = 100,\;\tau = 0.005,
\end{equation}
so that low-opacity splats (small $\alpha$) are perturbed more strongly than high-opacity ones.

\noindent\emph{Why this preserves appearance in expectation.}
Let $\mathbf{C}(\mathbf{u};\mathbf{X})$ denote the rendered color at pixel $\mathbf{u}$ given centers
$\mathbf{X}\!=\!\{\mathbf{x}_i\}$, and let
$\mathbf{X}' = \{\mathbf{x}_i'\}$ be the perturbed centers.
A second-order Taylor expansion of $\mathbf{C}$ around $\mathbf{X}$ yields
\begin{align}
\mathbb{E}_{\boldsymbol{\xi}}\big[\mathbf{C}(\mathbf{u};\mathbf{X}')\big]
&\approx \mathbf{C}(\mathbf{u};\mathbf{X})
\;+\;
\sum_i
\nabla_{\mathbf{x}_i}\mathbf{C}(\mathbf{u};\mathbf{X})^\top
\mathbb{E}[\Delta \mathbf{x}_i]
\notag\\[2pt]
&\quad
+\;
\frac{1}{2}
\sum_i
\mathrm{tr}\!\Big(
  H_i(\mathbf{u})\,
  \mathrm{Cov}[\Delta \mathbf{x}_i]
\Big),
\label{eq:taylor}
\end{align}
where $H_i(\mathbf{u})$ is the Hessian of $\mathbf{C}(\mathbf{u};\mathbf{X})$ \wrt $\mathbf{x}_i$.

From Eq.~\eqref{eq:noise}, we have $\mathbb{E}[\Delta \mathbf{x}_i] = \mathbf{0}$ and, using that $\Sigma_i$ is symmetric,
\begin{equation}
\mathrm{Cov}[\Delta \mathbf{x}_i]
=
\eta^2 w(\alpha_i)^2
\Sigma_i
\mathrm{Cov}[\boldsymbol{\xi}_i]
\Sigma_i^\top
=
\eta^2 w(\alpha_i)^2 \Sigma_i^2.
\end{equation}
Thus, the \emph{first-order} term in Eq.~\eqref{eq:taylor} vanishes, and the remaining \emph{second-order} bias is controlled by
$\eta^2 w(\alpha_i)^2 \Sigma_i^2$, \ie, it is (i) small for small $\eta$ and (ii) concentrated on low-opacity splats due to $w(\alpha_i)$.

\boldparagraph{Immature-manifold perturbation.}
RAIN-GS~\cite{jung2024relaxing} shows that 3DGS optimization follows a coarse-to-fine trajectory: early stages emphasize low-frequency structure, while finer details appear as covariances shrink. We mimic an earlier parameter state by inflating per-splat scales:
\begin{equation}
\textstyle
\mathbf{s}_i' \!=\! \mathbf{s}_i \odot \Big( 1 \!+\! (\beta-1)\,\omega(\mathbf{s}_i)\!\Big), 
\; \beta \sim \mathcal{U}[\beta_{\min},\beta_{\max}], \; \beta\!\geq\!1,
\end{equation}
where $\odot$ denotes element-wise multiplication, and $\omega(\mathbf{s}_i)\!\in\![0,1]$ is a scale-dependent weight that prioritizes small (high-frequency) splats. We define
\begin{equation}
\bar s_i \;=\; \tfrac{1}{3}\sum_{k=1}^3 s_{ik},
\qquad
\omega(\mathbf{s}_i)
=
\exp\!\Big(
  -\,\frac{\bar s_i}{\mathrm{median}_j\,\bar s_j}
\Big),
\end{equation}
so that splats with below-median size (small $\bar s_i$) receive larger inflation factors.

The induced covariance becomes
\begin{equation}
\Sigma_i'
=
R(\mathbf{q}_i)\,\mathrm{diag}((\mathbf{s}_i')^2)\,R(\mathbf{q}_i)^\top,
\end{equation}
and under local projection, the 2D covariance satisfies
\begin{equation}
\Sigma_i^{2\mathrm{D}}
\;\approx\;
J_i \Sigma_i J_i^\top,
\end{equation}
where $J_i$ is the Jacobian of the 3D$\to$2D projection at $\mathbf{x}_i$. 
A 2D Gaussian with covariance $\Sigma^{2\mathrm{D}}$ acts locally as a low-pass filter with Fourier-domain transfer function
\begin{equation}
\exp\!\Big(
  -\tfrac{1}{2} \boldsymbol{\kappa}^\top
  \Sigma^{2\mathrm{D}}
  \boldsymbol{\kappa}
\Big),
\end{equation}
for spatial frequency $\boldsymbol{\kappa}$. Inflating the scales from $\mathbf{s}_i$ to $\mathbf{s}_i'$ therefore increases the local 2D covariance from $\Sigma_i^{2\mathrm{D}}$ to $\Sigma_i^{\prime 2\mathrm{D}}$, which is equivalent to applying an additional Gaussian blur with covariance
\begin{equation}
\Delta \Sigma_i^{2\mathrm{D}}
=
\Sigma_i^{\prime 2\mathrm{D}} - \Sigma_i^{2\mathrm{D}}.
\end{equation}
This extra blur yields stronger high-frequency suppression while preserving coarse structure, which targets precisely the ``immature'' yet plausible states encountered early in 3DGS optimization.

\section{Pretraining Data}
\label{supp:data}

\subsection{Processing of Teacher Pseudo-Labels}
We employ three 2D teachers during \ours pretraining: \texttt{SigLIP2-so400m-p16-512}, \texttt{DINOv3-ViT-L/16}, and \texttt{PE-Spatial-L14-448}. Their dense feature maps have dimensionality 1152 (SigLIP2~\cite{tschannen2025siglip}) and 1024 (DINOv3~\cite{simeoni2025dinov3}, PE-Spatial~\cite{bolya2025perception}), respectively. For the selected RGB frames used in 3DGS optimization, we run all teachers offline and obtain the corresponding 2D feature maps. The input image resolutions and output feature map sizes for each teacher on each data source are summarized in \cref{tab:stats_teachers_res}.

\begin{table}[ht]
\centering
\scriptsize
\setlength{\tabcolsep}{2.0pt}
\renewcommand{\arraystretch}{1.05}

\begin{tabularx}{\linewidth}{@{}L{14mm} *{6}{C}}
\toprule
\multirow{2}{14mm}[-0.7ex]{\raggedright Input/Output \\Resolution}
  & \multicolumn{2}{c}{DINOv3}
  & \multicolumn{2}{c}{SigLIP2}
& \multicolumn{2}{c}{PE-Spatial} \\
\cmidrule(lr){2-3}\cmidrule(lr){4-5}\cmidrule(lr){6-7}
& input & output
& input & output  & input & output  \\
\cmidrule{1-7}

ScanNet
& 1296*968 & 81*60 & 640*480 & 616*456 & 640*480 & 45*34  \\
ScanNet++
& 1752*1168 & 109*73 & 876*584 & 876*584 & 876*584 & 62*41  \\
\matt
& 2560*2048 & 160*128 & 640*512 & 640*512 & 640*512 & 45*36  \\

\bottomrule
\end{tabularx}

\caption{\textbf{Input Image and Output Feature Map Size During 2D Feature Map Collection for Teachers.}}
\label{tab:stats_teachers_res}
\end{table}

For SigLIP2 and PE-Spatial, we use the same input resolutions as those used during 3DGS optimization. In contrast, DINOv3, due to its high-resolution-adapted training, is applied with high-resolution inputs: on ScanNet and ScanNet++ we use the original high-resolution RGB images, and on \matt we use upsampled images. SigLIP2 feature maps are obtained jointly with SAM2~\cite{ravi2024sam} inference, following Algorithm 1 in~\cite{li2025scenesplat}. The algorithm outputs feature maps with the same spatial size as its input images; the only exception is ScanNet, where we crop image borders, resulting in slightly smaller feature maps.

To obtain pseudo-labels on the Gaussians for each teacher, we lift the 2D teacher feature maps to the 3DGS, with the feature map spatial resolution being bilinearly interpolated to the image size during 3DGS optimization. For SigLIP2, we adapt the Occam’s LGS pipeline~\cite{cheng2024occam}, projecting each Gaussian into all views with valid features and aggregating the corresponding per-pixel embeddings into a single embedding per Gaussian based on rasterization weights. For DINOv3 and PE-Spatial, we use the adapted LUDVIG~\cite{marrie2025ludvig} codebase to perform uplifting, yielding 1024-D pseudo-labels per Gaussian for these two teachers. These per-Gaussian pseudo-labels form the supervision targets used in our multi-teacher pretraining framework.

\subsection{Update to SceneSplat-7K Data}

We largely use the 3DGS scenes collected in the SceneSplat-7K~\cite{li2025scenesplat} dataset. The only modification is to the Matterport3D~\cite{chang2017matterport3d} subset, where we observed limited 3DGS quality in its release. SceneSplat-7K is built from the \texttt{region\_segmentations} version provided by Matterport3D, and the original 3DGS optimization was performed independently for each region. Because RGB images were also split per region, each 3DGS was trained with only a small set of images, resulting in reduced multi-view coverage and degraded reconstruction quality.

\begin{figure}[ht]
    \centering
    \includegraphics[width=\linewidth]{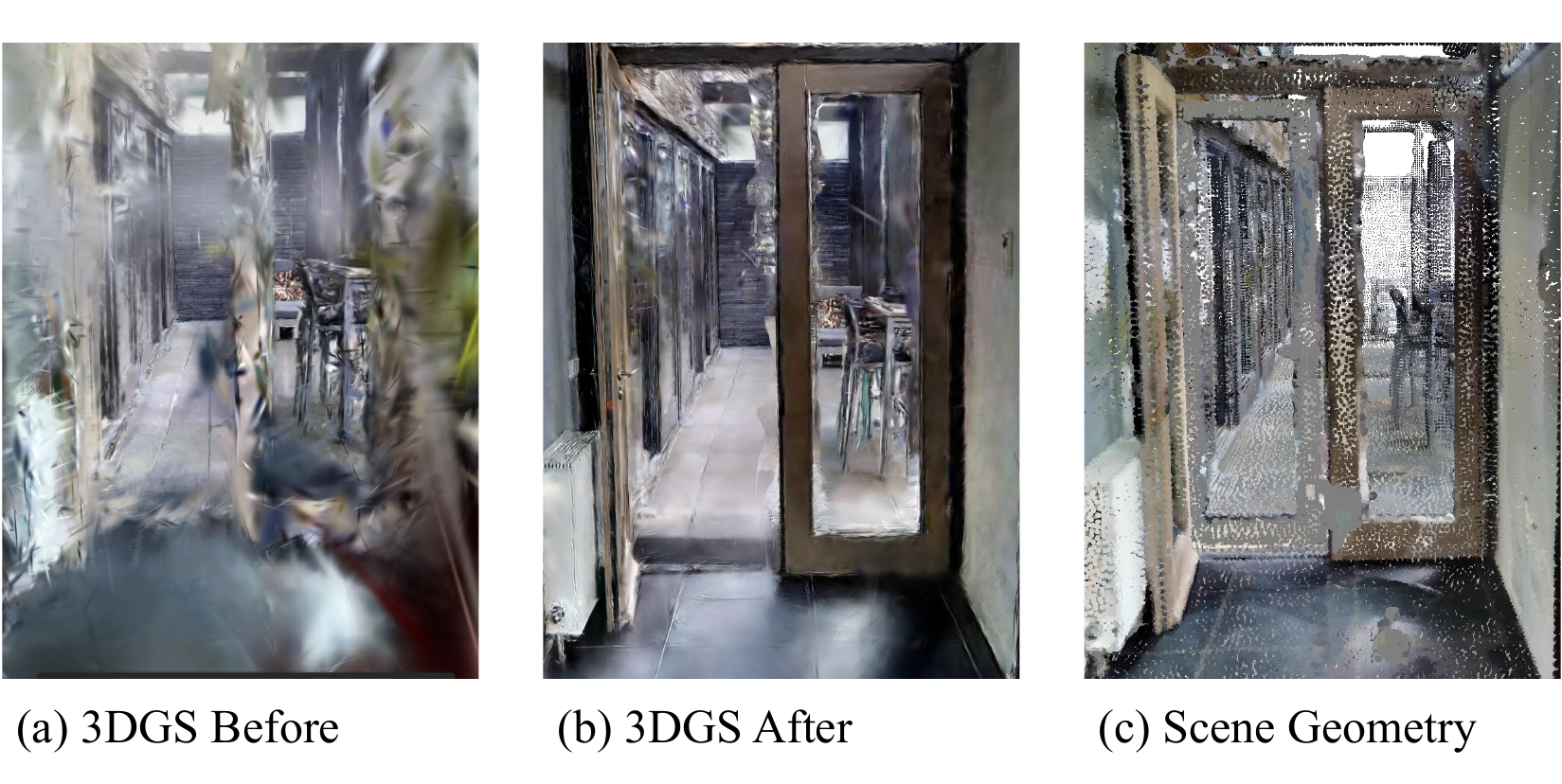}
    \vspace{-6mm}
    \caption{\textbf{Rendering Example of Updated \matt 3DGS in SceneSplat-7K.} }
    \label{fig:supp_data_update}
\end{figure}

\begin{figure}[ht]
    \centering
    \includegraphics[width=\linewidth]{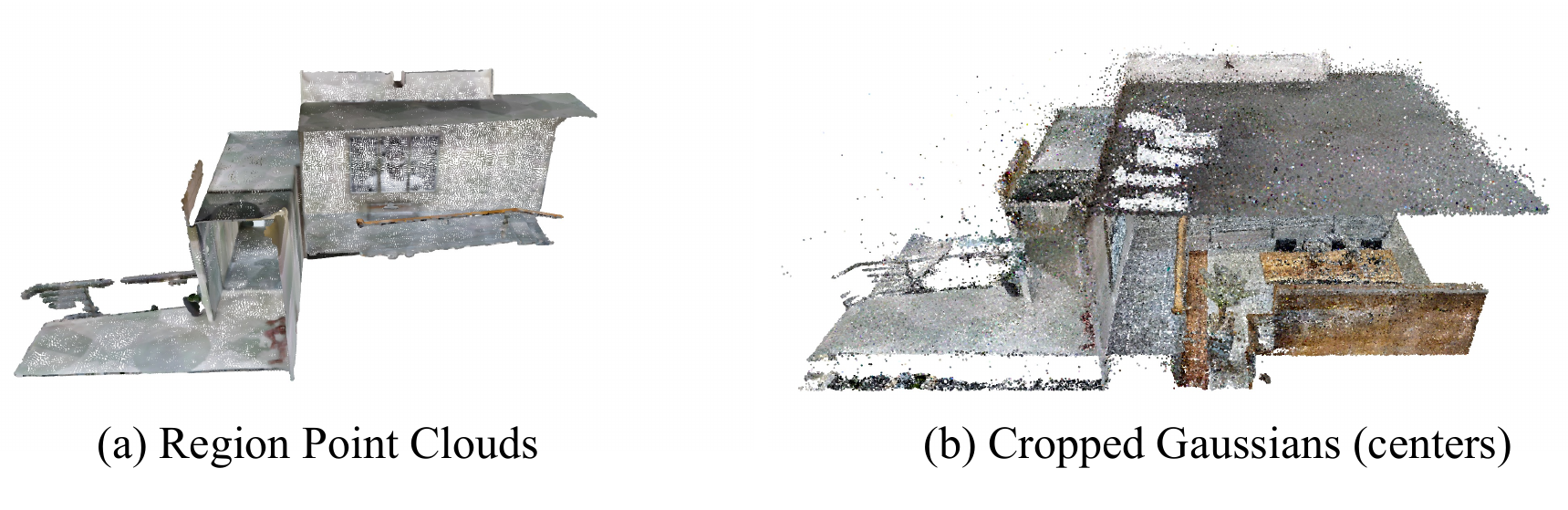}
    \vspace{-6mm}
    \caption{\textbf{Example of \matt Region Point Clouds and the Resulting Cropped Gaussians.} }
    \label{fig:mp3d_region_crop}
\end{figure}

In this work, we re-process the Matterport3D subset by first optimizing a single 3DGS scene per house using all available views, and then obtaining region-level 3DGS via cropping. Concretely, for each region we compute the axis-aligned bounding box of the corresponding region point clouds and enlarge it by $0.25\,\mathrm{m}$ in all directions before cropping the house-level Gaussians. This simple change yields noticeably better rendering quality for Matterport3D regions, as shown in~\cref{fig:supp_data_update}. An example of the original region point clouds and the resulting cropped Gaussians is shown in~\cref{fig:mp3d_region_crop}.

\subsection{Processing of \interior Data}
\label{supp:data_interior}

\interior~\cite{InteriorGS2025} contains 1{,}000 indoor scenes in Gaussian splats. Each scene provides instance-level semantic labels, instance IDs, 3D bounding boxes, and an occupancy map. We process \interior into a dataset suitable for our downstream benchmarks, including four main steps: semantic class mapping, assigning semantic and instance labels to Gaussians, outlier removal, and defining splits.

\boldparagraph{Semantic class mapping.}
The original annotation uses 755 fine-grained semantic categories, many of which are overly specific. We group and map these into 72 classes aimed at semantic clarity and hierarchy, while preserving privacy. Concretely, we remove any labels related to people, discard ambiguous tags such as \emph{"misc","set","object"} by mapping them to the ignore label ($-1$), and merge long-tail categories into broader concepts while keeping frequently occurring object types separate. The resulting 72 mapped classes each appear in at least 10 instances and together cover over 83.5\% of all labeled instances in \interior.

The mapped classes are summarized as follows:

\begin{itemize}
  \item Structural \& fixtures: bathtub, ceiling, ceiling light, column, countertop, door, faucet, floor, lamp, mirror, shower, sink, staircase, toilet, ventilation and heating, wall, window, window accessory.
  
  \item Furniture \& appliances: bed, bedding, bench, bookshelf, cabinet, chair, clock, computer, computer peripheral, cooktop, decor, desk, dishwasher, dresser, drawer, fan, floor covering, mattress, microwave, monitor, oven, plant, range hood, refrigerator, shelving, sofa, stool, stove, table, television, trash can, vase, wall art, wardrobe, washing machine.
  
  \item Goods \& supplies (coarse): bakery, beverage, books, container, cookware, electronics accessory, medicine, packaged food, paper goods, produce, snacks and candy, stationery, tableware, tobacco, toiletries, towel, toy, utensils, wine.
\end{itemize}

\boldparagraph{Assigning semantic and instance labels to Gaussians.}
In the original \interior annotations, semantics and instance IDs are attached to 3D bounding boxes, not directly to Gaussians. Simply labeling every Gaussian inside a box is problematic when boxes overlap, as this causes label “bleeding” between neighboring objects (\eg, a table partially swallowed by a chair’s bounding box). To make the assignment more accurate, we label each box using a connected-components heuristic.

\begin{figure}[ht]
    \centering
    \includegraphics[width=0.9\linewidth]{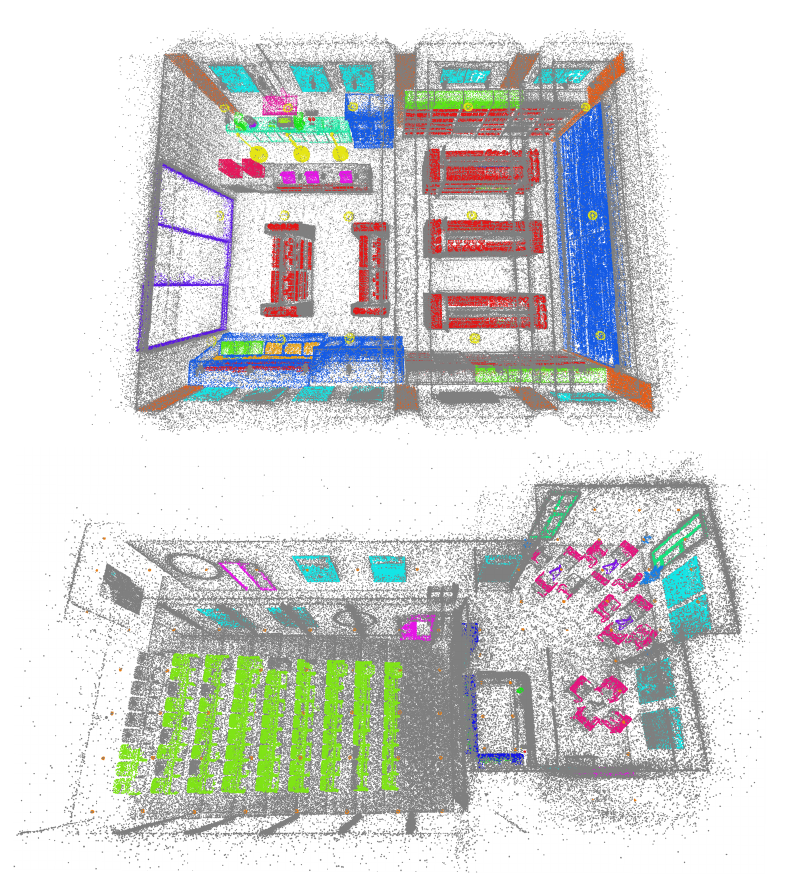}
    \caption{\textbf{Semantic Labels Visualization (Gaussian centers) of \interior Scenes.} Different classes are in distinct colors with the ignore label in dark grey.}
    \label{fig:supp_interiorgs}
\end{figure}

For each annotated bounding box, we first obtain Gaussians whose centers lie inside the box and voxelize these points on a sparse 3D grid. We then compute connected components over the occupied voxels and retain only the largest connected component (LCC). Semantic and instance labels from the box are assigned only to Gaussians whose voxels belong to this LCC, which suppresses overlaps with adjacent bounding boxes while preserving the target instance. If the box contains fewer than two occupied voxels, we skip this bounding box label. After this step, each Gaussian is associated with at most one instance ID and a semantic label from the 72-class taxonomy. Examples of assigned semantic labels are visualized in \cref{fig:supp_interiorgs}.

\boldparagraph{Outlier removal.}
Some \interior scenes contain a small number of Gaussians located far from the scene contents. We remove such outliers using a simple quantile-based filter on the Gaussian centers. For each scene and each axis (x, y, z), we compute the $q$ and $(1-q)$ percentiles with $q = 10^{-3}$, giving an interval $[\text{lower}, \text{upper}]$. We then enlarge this interval by 5\% of its span on each side and keep only Gaussians whose centers lie within these expanded bounds on all three axes. In other words, we calculate
\begin{equation}
\text{span} = \text{upper} - \text{lower}, \quad
\text{buffer} = 0.05 \cdot \text{span}.
\end{equation}
The resulting valid interval is $[\text{lower} - \text{buffer}, \text{upper} + \text{buffer}]$. This procedure effectively removes outlier Gaussians that are far away, accounting for $0.20\%$ of the total splats.

\boldparagraph{Dataset splits.}
Finally, we define a simple scene-level split of \interior for benchmarking: scenes \texttt{001-100} are used as the test set, \texttt{101-200} as the validation set, and \texttt{201-1000} as the training set. Note that \interior is \emph{not} included in the pretraining data of \ours; it is reserved entirely for zero-shot evaluation, downstream probing, and rendering-based adaptation.

\section{Implementation Details}
\label{supp:impl}

\subsection{Open-Vocabulary Segmentation}
\label{supp:impl_open_vocab}
In the main paper Tabs.~1 and 2, when running \ours and~\cite{li2025scenesplat} for open-vocabulary semantic and instance segmentation on the \scannet, \ppv2, and \matt benchmarks, predictions are first obtained at the Gaussians' centers and then propagated to the ground-truth labeling locations through nearest-neighbor voting. When running the point-cloud baseline~\cite{lee2025mosaic3d} on \interior, we compared two evaluation inputs: \textit{(i)} Gaussians' centers, colors, and estimated normals; and \textit{(ii)} point clouds generated by sampling viewpoints, fusing rendered depth maps from test scenes, and interpolating colors from the Gaussians. We choose approach \textit{(i)} because it leads to better results for the baseline.

\subsection{Language Model-Based Question Answering}
We follow GaussianVLM~\cite{halacheva2025gaussianvlm} to conduct VLM experiments with our Chorus encoder. We represent each 3D scene using $40$k randomly sampled Gaussians from SceneSplat~\cite{li2025scenesplat,ma2025scenesplat++}. For the language model, we adopt OPT-1.3B~\cite{zhang2022opt}, following the LL3DA training protocol. The LLM is loaded in \texttt{float16} for memory efficiency and finetuned using LoRA. Our training procedure adheres to the standard protocol: $5$ epochs of alignment followed by $32$ epochs of instruction tuning. Training completes in under one day on two H200 GPUs. Additionally, we pretrain our task-guided sparsifier on the object-captioning task for $5$ epochs. We employ the AdamW optimizer with a weight decay of $0.1$ and a cosine-annealing learning-rate schedule, decaying from $10^{-4}$ to $10^{-6}$. Evaluation is performed every $8$ epochs. 

For the connector, we preserve the overall sparsifier architecture and only modify the task-guided sparsifier: we reduce it to a single attention block and feed it only the final-layer Chorus features as input, leaving all other components unchanged.
\paragraph{Evaluation Metrics.}
For a predicted sentence $\hat{y}$ and reference sentence $y$, we use the following metrics; all scores are averaged over the evaluation set.

\begin{itemize}
    \item \textbf{EM1 (Top-1 Exact Match).}  
    This metric measures the fraction of predictions that exactly match the reference answer (after normalization such as lowercasing and stripping punctuation).  
    For $N$ samples with predictions $\{\hat{y}_i\}_{i=1}^N$ and references $\{y_i\}_{i=1}^N$, we compute
    \begin{equation}
        \text{EM}_1 = \frac{1}{N} \sum_{i=1}^{N} \mathbb{I}\big[\hat{y}_i = y_i\big],
    \end{equation}
    where $\mathbb{I}[\cdot]$ is the indicator function.

    \item \textbf{M (METEOR)}~\cite{banerjee2005meteor}.  
    METEOR computes a unigram alignment between $\hat{y}$ and $y$ (accounting for stem, synonym, and paraphrase matches), and then derives precision and recall over matched unigrams:
    \begin{equation}
        P = \frac{\text{\# matched unigrams}}{\text{\# unigrams in } \hat{y}}, 
        \qquad
        R = \frac{\text{\# matched unigrams}}{\text{\# unigrams in } y}.
    \end{equation}
    A recall-weighted F-score is then defined as
    \begin{equation}
        F_{\beta} = \frac{(1+\beta^2)\, P R}{\beta^2 P + R},
    \end{equation}
    where $\beta > 1$ emphasizes recall (we use the default METEOR setting).  
    Finally, a fragmentation penalty is applied based on the number of contiguous matched chunks $ch$ and the total number of matched unigrams $m$:
    \begin{equation}
        \text{Penalty} = \gamma \left(\frac{ch}{m}\right)^{\theta},
        \qquad
        \text{METEOR} = (1 - \text{Penalty}) \cdot F_{\beta},
    \end{equation}
    with $\gamma$ and $\theta$ as method-specific hyperparameters.

    \item \textbf{R (ROUGE)}~\cite{lin2004rouge}.  
    We report ROUGE-L, which is based on the length of the longest common subsequence (LCS) between $\hat{y}$ and $y$.  
    Let $\text{LCS}(\hat{y}, y)$ denote the LCS length, and let $|\hat{y}|$ and $|y|$ be the sequence lengths. We define
    \begin{equation}
        P_L = \frac{\text{LCS}(\hat{y}, y)}{|\hat{y}|}, 
        \qquad
        R_L = \frac{\text{LCS}(\hat{y}, y)}{|y|},
    \end{equation}
    and compute an $F$-score:
    \begin{equation}
        F^{\text{ROUGE-L}}_{\beta} = \frac{(1+\beta^2)\, P_L R_L}{\beta^2 P_L + R_L},
    \end{equation}
    where $\beta$ controls the relative weighting of recall (we use the standard ROUGE-L configuration).

    \item \textbf{Sim (Sentence Similarity)}~\cite{reimers-2019-sentence-bert}.  
    To capture semantic similarity beyond surface-level overlap, we encode $\hat{y}$ and $y$ with Sentence-BERT to obtain embeddings $\mathbf{e}_{\hat{y}}, \mathbf{e}_{y} \in \mathbb{R}^d$.  
    We then compute cosine similarity:
    \begin{equation}
        \text{Sim}(\hat{y}, y) 
        = \cos\big(\mathbf{e}_{\hat{y}}, \mathbf{e}_{y}\big)
        = \frac{\mathbf{e}_{\hat{y}}^{\top}\mathbf{e}_{y}}
        {\|\mathbf{e}_{\hat{y}}\|_2 \,\|\mathbf{e}_{y}\|_2}.
    \end{equation}
\end{itemize}

\subsection{Chorus on Point-Cloud Tasks}
For all results reported in the main paper Tabs.~4, 5, and 6 on Semantic Segmentation Probing \& Finetuning, the ScanNet Data-Efficient Benchmark, and Instance Segmentation Probing and Finetuning, we conduct the experiments using data epochs of 400. We reproduced~\cite{wu2025sonata} using the official checkpoint and report results for~\cite{wu2024ppt,wu2023msc} from~\cite{zhang2025concerto}. During Semantic Segmentation Probing \& Finetuning on \interior, we use the standard \ours encoder\gs that takes Gaussian parameters as input, and for the point-cloud baseline~\cite{wu2025sonata}, similarly to~\cref{supp:impl_open_vocab}, we have chosen the input with Gaussians' centers, colors, and estimated normals, as it leads to better results for the baseline. For all other benchmarks, we use the \ours variant\pts that takes point-cloud inputs.

\subsection{Rendering-Based Adaptation}

We present the 2D statistical results for the final selected scenes in \cref{tab:2d_data_summary}, which were filtered based on two criteria: the minimum number of available poses must be greater than 10, and scenes with too many splats (where the minimum visible Gaussian count exceeds $819\text{k}$) are ignored. Crucially, the box augmentation strategy significantly increases the number of splats. This effectively increases the number of output features supervised by the 2D teacher models.

We use a base resolution of $480\times 640$ for image rendering, as outlined in step 3 of \cref{alg:2d_adaption_train}. To generate a larger and higher-resolution feature map, we interpolate the images before feeding them into the 2D teacher model (rendering directly to high-resolution images would be slightly slower). In the resolution experiments, we interpolate images up to $960\times 1280$ to generate the $60\times 80$ DINOv3 feature map. By default, the predicted feature map is rendered to a resolution of $120\times 160$ (step 5 in \cref{alg:2d_adaption_train}), which is also the dimension used for the interpolation of the teacher features.

\begin{table}[h]
    \centering 
    \scriptsize
    \begin{tabularx}{\linewidth}{L{15mm}L{20mm}XX} 
        \toprule
        \textbf{Avg. Frame Number} & \textbf{Avg. per-frame visible Gaussians} & \textbf{+ box cropping} & \textbf{+ overlapping pair merge} \\
        \midrule
        37 & 126k (16\%) & 348k (43\%)  & 419k (52\%) \\
        \bottomrule
    \end{tabularx}
    \caption{\textbf{Pose Sampling Statistics on \interior.} In 582 selected scenes, we report the average number of visible Gaussians in each frame and their ratio relative to the total number of Gaussians in the scene.}
    \label{tab:2d_data_summary}
\end{table}

\section{Additional Experiment Results}
\label{supp:exp}

\subsection{Language Teacher Ablation}
\label{supp:exp_ablation}

\begin{table}[ht]
\centering
\footnotesize
\setlength{\tabcolsep}{2.0pt}
\renewcommand{\arraystretch}{1.05}

\begin{tabularx}{\linewidth}{@{}L{18mm} ccc *{4}{C}}
\toprule
\multirow{2}{20mm}[-0.7ex]{\raggedright Training\\source}
  & \multicolumn{3}{c}{Teachers}
  & \multicolumn{2}{c}{Linear Probing}
& \multicolumn{2}{c}{Decoder Probing} \\
\cmidrule(lr){2-4} \cmidrule(lr){5-8}
& DINO & Lang & PE
& mIoU & mAcc & mIoU & mAcc \\

\midrule

\multirow{2}{*}{ScanNet}
& \checkmark & - & - & 23.7 & 34.2 & 26.7 & 37.6 \\
& \checkmark & \checkmark & - & \cellfirst{24.7} & \cellfirst{35.6} & \cellfirst{27.5} & \cellfirst{38.2} \\
\cmidrule{1-8}

\multirow{4}{*}{ScanNet++ v2}
& \checkmark & - & - & 24.6 & 35.2 & 26.2 & 37.4 \\
& - & \checkmark & - & 24.4 & 35.1 & 27.4 & 40.0 \\
& \checkmark & \checkmark & - & 25.1 & 36.7 & \cellfirst{28.9} & 40.4 \\
& \checkmark & \checkmark & \checkmark & \cellfirst{25.6} & \cellfirst{37.3} & 28.3 & \cellfirst{40.6} \\

\bottomrule
\end{tabularx}

\caption{\textbf{Language Teacher Ablation on InteriorGS.}
We report linear probing and decoder probing results for the corresponding pretrained \ours encoder.}
\label{tab:ablation_lang_teacher}
\end{table}

In the main paper Tab.~9, we ablate the teachers by fixing the language teacher (SigLIP~\cite{tschannen2025siglip}) and then adding DINO~\cite{simeoni2025dinov3}, followed by PE-Spatial~\cite{bolya2025perception}, using zero-shot segmentation performance on \interior. \cref{tab:ablation_lang_teacher} presents another view where we start with the DINO teacher, then add SigLIP and PE-Spatial. In this setting, we compare the probing performance on the \interior dataset, as an encoder trained with the DINO teacher alone does not support zero-shot segmentation. The results are mostly consistent with Tab.~9: incorporating both the SigLIP and PE-Spatial teachers yields an improvement in probing accuracy, verifying that language-aligned and object-aware cues are complementary to the generalist features from DINO.

\begin{figure}[h]
    \centering
    \includegraphics[width=0.9\linewidth]{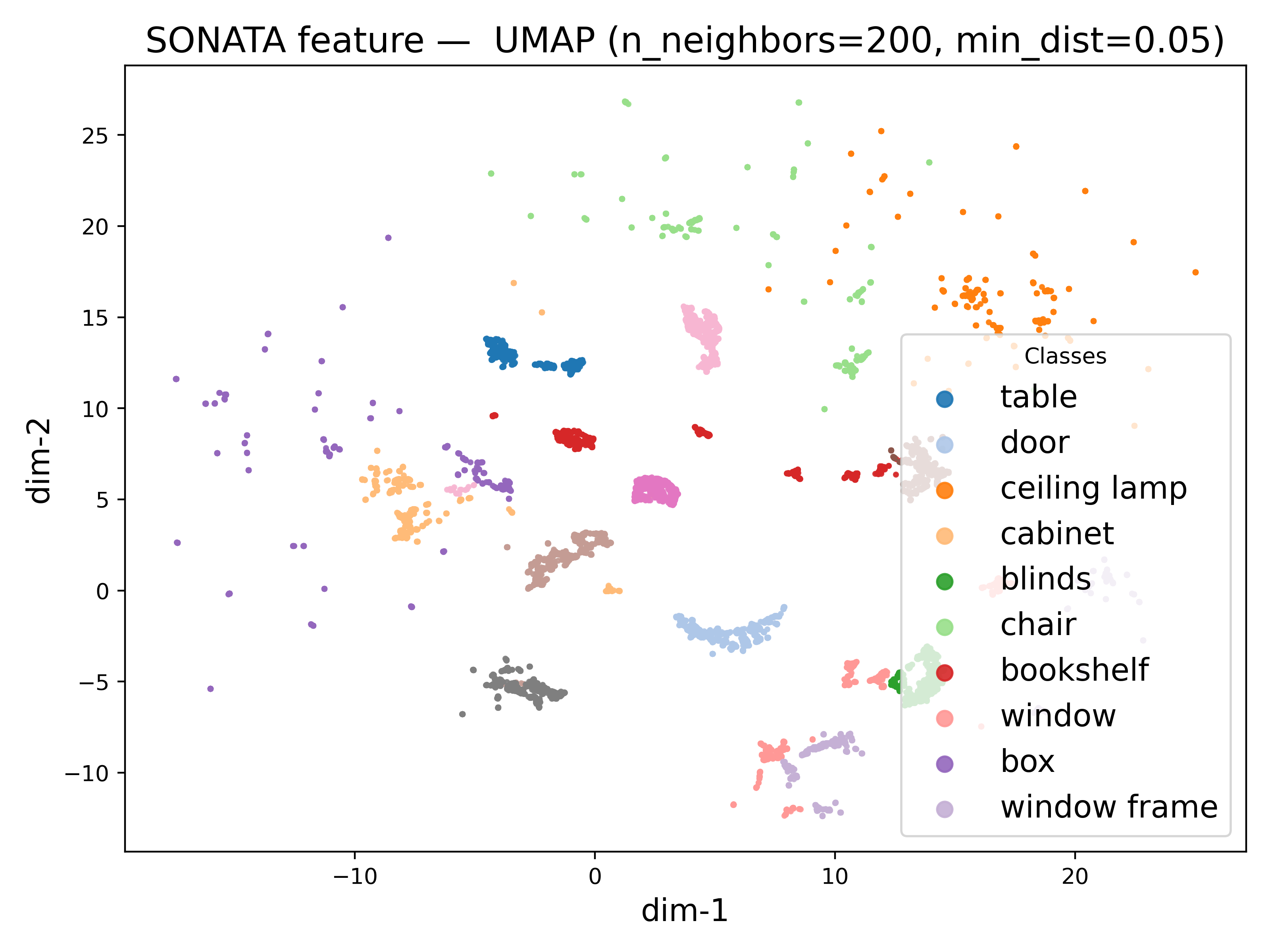}
    \vspace{1em}
    \includegraphics[width=0.9\linewidth]{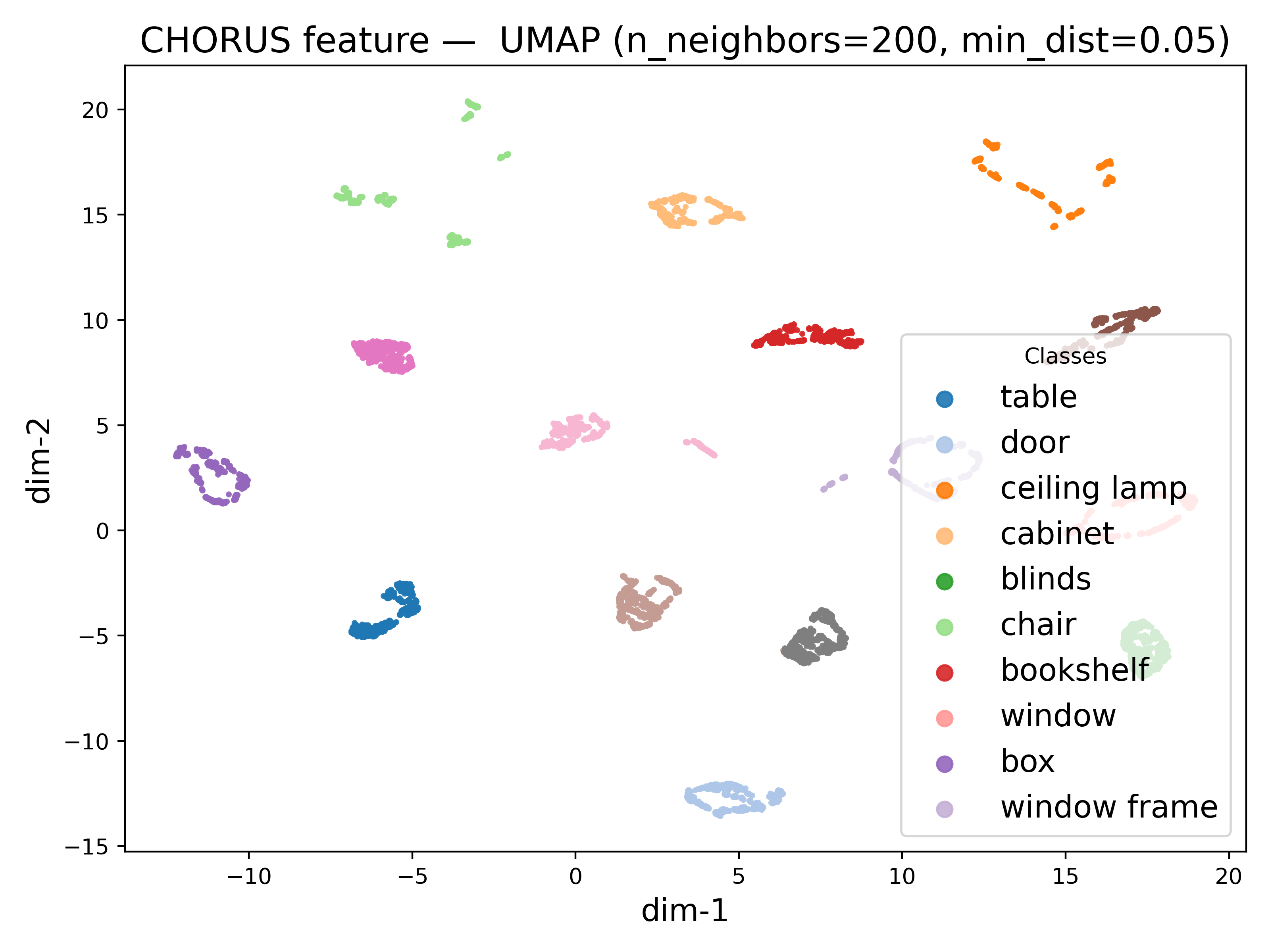}
    \caption{\textbf{UMAP Feature Analysis.} We select 15 semantic classes, sample one instance per class from 10 scenes in the \ppv2 Val split, and then project per-point encoded features from each method using UMAP. \ours leads to more compact and well-separated clusters.}
    \label{fig:umap_feat}
\end{figure}

\begin{figure*}[t]
    \centering
    \includegraphics[width=0.85\textwidth]{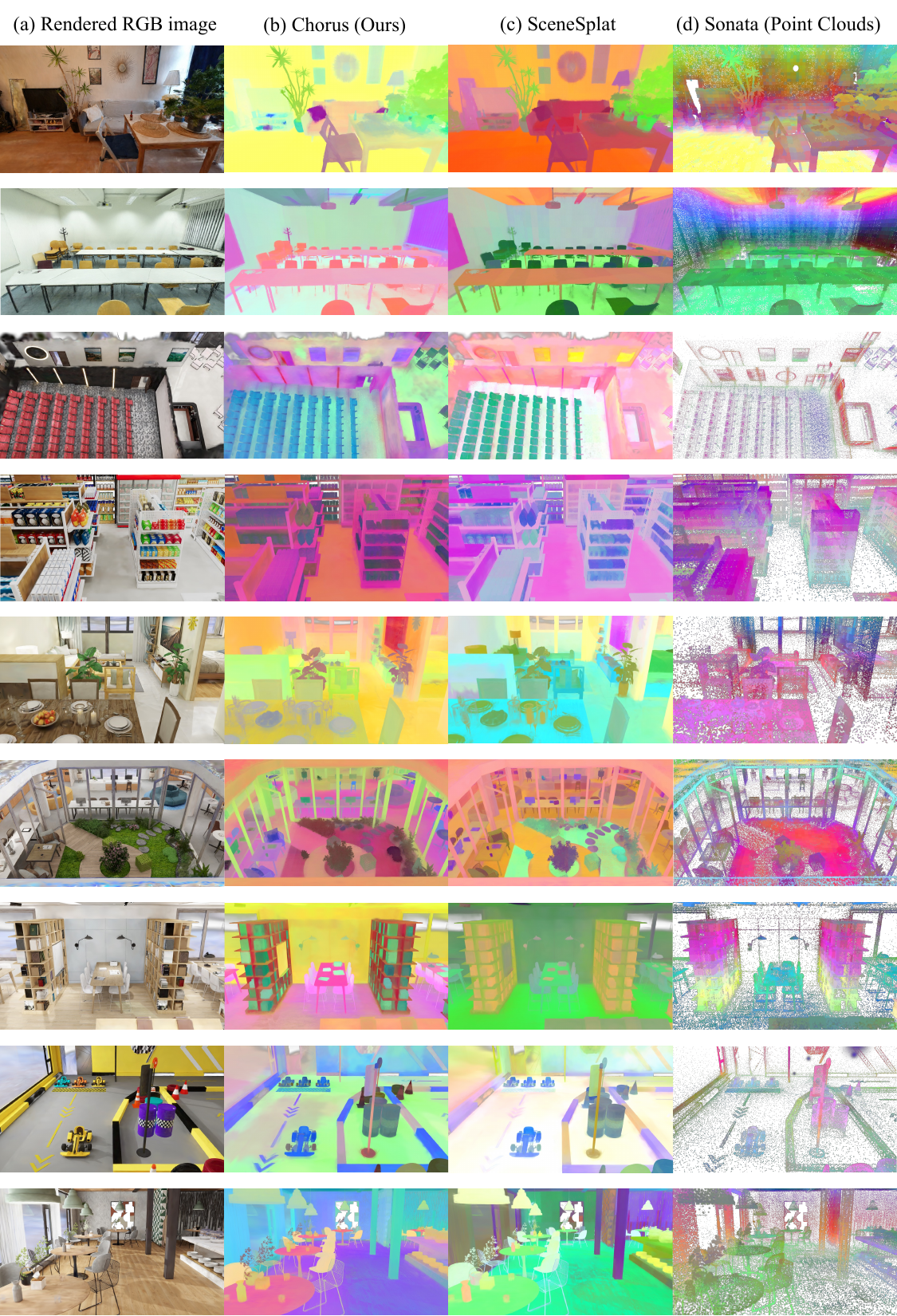}
    \caption{\textbf{Inference Feature PCA Comparison} (best viewed zoomed in).}
    \label{fig:supp_feature_pca}
\end{figure*}

\subsection{UMAP Feature Analysis}
\label{supp:umap_feat}

To further support our hypothesis on why \ours works well on point clouds, we compare the Chorus variant that operates directly on point clouds with the Sonata~\cite{wu2025sonata} encoder trained with the self-supervised scheme, by visualizing the encoded features with UMAP~\cite{mcinnes2018umap}. We select 15 semantic classes, sample one instance per class, and subsample at most 1000 points per instance from 10 scenes in the \ppv2 Val split. We then project per-point encoded features from each method using UMAP with $n_\text{neighbors}=200$ and $\text{min\_dist}=0.05$. We choose UMAP instead of t-SNE because it better preserves global neighborhood structure, making inter-cluster distances more interpretable.

As shown in \cref{fig:umap_feat}, Chorus produces noticeably more compact and well-separated clusters than Sonata, especially for classes such as \emph{ceiling lamp} and \emph{box}, where Sonata exhibits more scatter and overlap with other categories. The tight within-instance clusters and clear inter-class margins indicate that Chorus learns semantically consistent and geometry-aware embeddings even when only Gaussians' centers, colors, and normals are used as inputs during pretraining. These observations align with our retrieval results in the main paper ablation study: 3DGS-based pretraining behaves like a strong augmentation that stabilizes point-cloud features, while multi-teacher supervision encourages a more structured feature space that transfers effectively to point clouds despite the inference distribution gap.

\subsection{Inference Feature PCA Visualization}
\label{supp:feature_pca}
We provide additional qualitative examples in~\cref{fig:supp_feature_pca} comparing the scene encodings of \ours, SceneSplat~\cite{li2025scenesplat}, and Sonata~\cite{wu2025sonata} on diverse scenes from the \ppv2 Val split and the \interior Test split. Compared to SceneSplat, \ours exhibits stronger semantic consistency; for example, instances of the semantic classes \emph{chair}, \emph{plant}, \emph{food package}, and \emph{book} are more coherently clustered. \ours also shows superior geometry awareness: even a stylized chair with many thin structures is correctly encoded, \ie, its PCA colors resemble those of other chairs, whereas SceneSplat fails to recognize it. Meanwhile, the encodings from Sonata exhibit noticeable grid artifacts and reduced semantic consistency. These qualitative trends are consistent with the performance gains of our Chorus encoder on 3DGS-native and point-cloud benchmarks reported in the main paper.

\subsection{2D Adaptation Feature PCA Visualization}

\begin{figure}[ht]
    \centering
    \centering
    \begin{subfigure}[b]{0.31\linewidth}
        \centering
        \includegraphics[width=\linewidth]{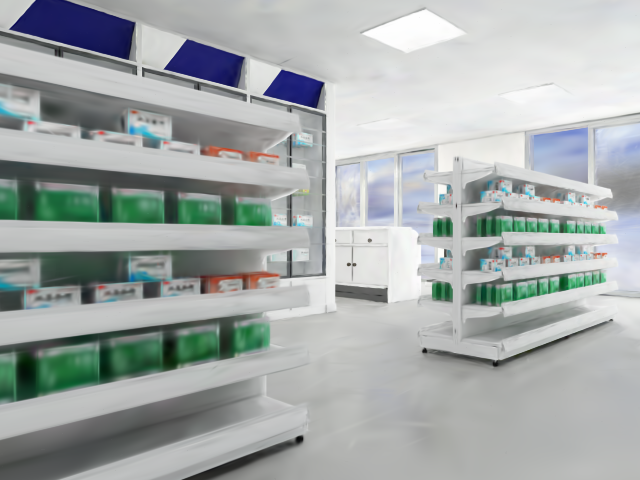}
        \caption{}
    \end{subfigure}
    \hfill
    \begin{subfigure}[b]{0.31\linewidth}
        \centering
        \includegraphics[width=\linewidth]{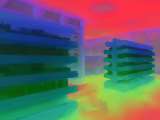}
        \caption{}
    \end{subfigure}
    \hfill
    \begin{subfigure}[b]{0.31\linewidth}
        \centering
        \includegraphics[width=\linewidth]{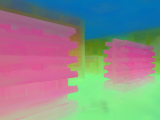}
        \caption{}
    \end{subfigure}

    \caption{\textbf{Inference Features of the \ours DINO Projector Head Before and After Rendering-Based Adaptation.} (a) Rendered RGB image. (b) PCA of rendered features before adaptation. (c) PCA of rendered features after adaptation.}
    \label{fig:after_2d_adaptation}
\end{figure}

\Cref{fig:after_2d_adaptation} visualizes the features of an unseen InteriorGS~\cite{InteriorGS2025} scene processed by \ours. The adaptation notably reduces inconsistency on the ground plane. However, the lamp and ceiling are not clearly distinguishable, and the edges are less distinct. This degradation is likely due to the use of a low-resolution feature map for supervision. As shown in the ablation in our main paper~\cref{fig:2D_adapt_ablation}, employing higher-resolution feature maps is preferred to improve adaptation performance.

\section{Impact Statement}
\label{supp:impact}
\ours advances 3D scene understanding by offering the first holistic, feed-forward 3DGS encoder pretrained with multi-teacher distillation. By unifying complementary signals from language-aligned, general-purpose, and object-aware 2D foundation models, \ours produces highly structured scene embeddings that generalize across tasks, datasets, and even modalities (3DGS and point clouds) with remarkable data efficiency. This capability enables more reliable open-vocabulary perception and more scalable scene understanding in applications like XR and embodied AI systems. Beyond in-distribution performance, our lightweight render-and-distill adaptation pipeline lowers the barrier for deploying the model to new environments without expensive 3D annotations. Overall, \ours contributes a practical and scalable route for the next generation of 3D scene encoders.

\end{document}